\documentclass[10pt,journal,compsoc]{IEEEtran}
\usepackage{ifpdf}
\ifpdf
\else
\fi

\ifCLASSOPTIONcompsoc
  \usepackage[nocompress]{cite}
\else
  \usepackage{cite}
\fi
\ifCLASSINFOpdf
   \usepackage[pdftex]{graphicx}
   \graphicspath{{../pdf/}{../jpeg/}}
   \DeclareGraphicsExtensions{.pdf,.jpeg,.png}
\else
   \usepackage[dvips]{graphicx}
   \graphicspath{{../eps/}}
   \DeclareGraphicsExtensions{.eps}
\fi
\usepackage{amsmath}
\usepackage{array}

\ifCLASSOPTIONcompsoc
 \usepackage[caption=false,font=footnotesize,labelfont=sf,textfont=sf]{subfig}
\else
 \usepackage[caption=false,font=footnotesize]{subfig}
\fi
\usepackage{fixltx2e}

\usepackage{stfloats}
\ifCLASSOPTIONcaptionsoff
 \usepackage[nomarkers]{endfloat}
\let\MYoriglatexcaption\caption
\renewcommand{\caption}[2][\relax]{\MYoriglatexcaption[#2]{#2}}
\fi
\let\MYorigsubfloat\subfloat
\renewcommand{\subfloat}[2][\relax]{\MYorigsubfloat[]{#2}}
\usepackage{xspace}
\makeatletter
\DeclareRobustCommand\onedot{\futurelet\@let@token\@onedot}
\def\@onedot{\ifx\@let@token.\else.\null\fi\xspace}

\def\eg{\emph{e.g}\onedot} 
\def\ie{\emph{i.e}\onedot} 
 
\def\etc{\emph{etc}\onedot} 
 
\def\etal{\emph{et al}\onedot}
\makeatother

\usepackage{enumitem}
\usepackage{caption}
\usepackage{times}
\usepackage{epsfig}
\usepackage{graphicx}
\usepackage{amsmath}
\usepackage{amssymb}
\usepackage{eso-pic}
\usepackage{helvet}
\usepackage{courier}
\usepackage{url}
\usepackage{graphicx}
\usepackage{mathrsfs}
\usepackage{multirow}
\usepackage{comment}
\usepackage{marvosym}
\usepackage{bbm}
\usepackage{enumerate}
\usepackage{soul}
\usepackage[utf8]{inputenc}
\usepackage{amsfonts}
\usepackage{nicefrac}
\usepackage{microtype}
\usepackage{booktabs} 
\usepackage{algorithm}
\usepackage{algorithmicx}
\usepackage{algpseudocode}
\usepackage{makecell}
\usepackage{multirow}

\usepackage{bbding}
\usepackage[numbers]{natbib}
\usepackage[breaklinks=true,bookmarks=false,colorlinks]{hyperref}

\newcommand{\eqn}[1]{Eq.~\eqref{#1}}
\newcommand{\sect}[1]{Section~\ref{#1}}
\newcommand{\tab}[1]{Table~\ref{#1}}
\newcommand{\fig}[1]{Fig.~\ref{#1}}
\newcommand{\alg}[1]{Alg.~\ref{#1}}
\newcommand{\myparagraph}[1]{\vspace{5pt} \noindent \textbf{#1}}

\algdef{SE}[DOWHILE]{Do}{doWhile}{\algorithmicdo}[1]{\algorithmicwhile\ #1}%

\begin{document}

\title{Out-of-Domain Human Mesh Reconstruction via Dynamic Bilevel Online Adaptation}

%
%
%
%

\author{Shanyan Guan,
        Jingwei Xu,
        Michelle Z. He,
        Yunbo~Wang,
        Bingbing Ni,
        Xiaokang Yang,~\IEEEmembership{Fellow,~IEEE}
\IEEEcompsocitemizethanks{
\IEEEcompsocthanksitem S. Guan, J. Xu, Y. Wang, B. Ni, and X. Yang are with MoE Key Lab of Artificial Intelligence, AI Institute, Shanghai Jiao Tong University, China.
\IEEEcompsocthanksitem M. Z. He is with Purdue University, USA. Work done in part at Shanghai Jiao Tong University.
\IEEEcompsocthanksitem Corresponding author: Y. Wang, yunbow@sjtu.edu.cn.
\IEEEcompsocthanksitem Project page: \url{https://sites.google.com/view/dynaboa}.
}
}

\markboth{IEEE Transactions on Pattern Analysis and Machine Intelligence,~Vol.~XX, No.~X, March~2021}%
{Guan \MakeLowercase{\textit{et al.}}: Out-of-Domain Human Mesh Reconstruction via Dynamic Bilevel Online Adaptation}

\IEEEtitleabstractindextext{%

\begin{abstract}
We consider a new problem of adapting a human mesh reconstruction model to out-of-domain streaming videos, where performance of existing SMPL-based models are significantly affected by the distribution shift represented by different camera parameters, bone lengths, backgrounds, and occlusions. We tackle this problem through online adaptation, gradually correcting the model bias during testing. There are two main challenges: First, the lack of 3D annotations increases the training difficulty and results in 3D ambiguities. Second, non-stationary data distribution makes it difficult to strike a balance between fitting regular frames and hard samples with severe occlusions or dramatic changes. To this end, we propose the Dynamic Bilevel Online Adaptation algorithm (DynaBOA). It first introduces the temporal constraints to compensate for the unavailable 3D annotations, and leverages a bilevel optimization procedure to address the conflicts between multi-objectives. DynaBOA provides additional 3D guidance by co-training with similar source examples retrieved efficiently despite the distribution shift. Furthermore, it can adaptively adjust the number of optimization steps on individual frames to fully fit hard samples and avoid overfitting regular frames. DynaBOA achieves state-of-the-art results on three out-of-domain human mesh reconstruction benchmarks.

\end{abstract}



}

\maketitle

\IEEEdisplaynontitleabstractindextext

%
\IEEEpeerreviewmaketitle

\section{Introduction}

\begin{figure}[t]
    \centering
    \includegraphics[width=\columnwidth]{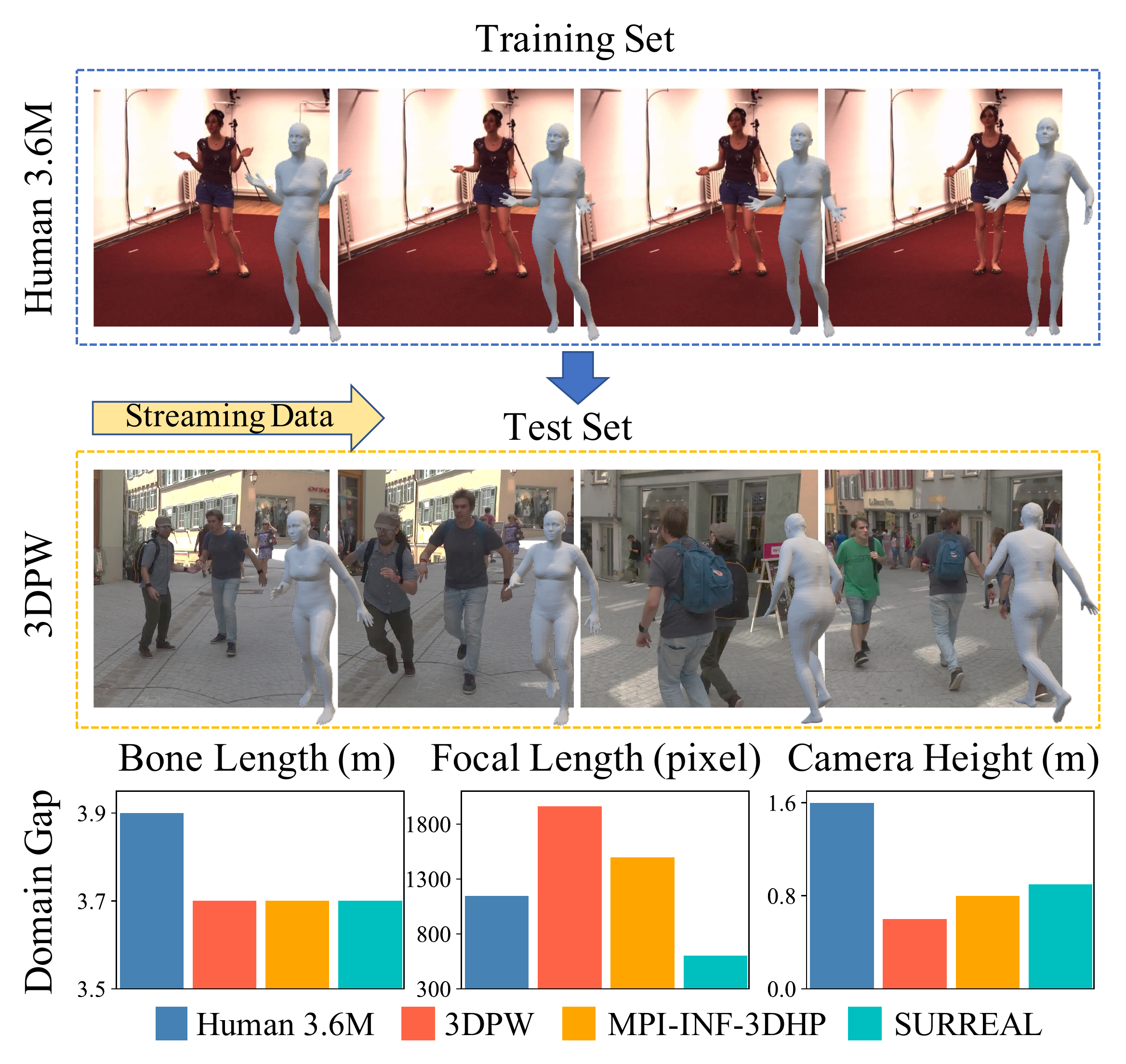}
    \caption{\textbf{Top:} Problem setup of learning to reconstruct human meshes from out-of-domain streaming videos. In terms of visual inputs, typical domain gaps include plain/crowded backgrounds, slight/heavy occlusions, simple/complex motions, etc.
    \textbf{Bottom:} Other domain gaps are reflected in the statistics of body skeleton and camera parameters. Bone length refers to the sum of the lengths between human joints, whose topology is shared across datasets.}
    \label{fig:problem}
    \vspace{-5pt}
\end{figure}

Human mesh reconstruction from streaming videos shows remarkable significance in real-world applications, \eg, sports performance analysis~\cite{rematas2018soccer,chen2021sportscap}, VR/AR~\cite{lin2016virtual,cha2018towards,weng2019photo} and human–computer interaction~\cite{zhang2020perceiving,guzov2021human,hassan2021populating}.
%
Most existing models assume that training and testing videos are identically distributed~\cite{kanazawa2018end,pavlakos2018learning,kolotouros2019learning,aksan2019structured,xu2019denserac,guler2019holopose,Moon_2020_ECCV_I2L-MeshNet,kocabas2020vibe}.
However, this assumption does not always hold in practice, as it is difficult to cover the richness of in-the-wild data during the training phase, so that there can be large distribution shifts between training and testing scenarios.
As shown in \fig{fig:problem}, typical distribution shifts are reflected in camera parameters, lengths of body bones, backgrounds, and occlusions.
Reconstructing human mesh from out-of-domain videos is under-explored and challenging, in which most of the advanced models, including SPIN \cite{kolotouros2019learning} and VIBE \cite{kocabas2020vibe}, underperform due to inadequate generalization ability.

In this paper, we study the problem of out-of-domain human mesh reconstruction from a new perspective, \ie, adapting the model online at test time before making the prediction.
The key insight is that the test samples, although being unlabeled, contain visual information that can be properly used to progressively correct the bias of the model against the distribution shift and thus improve its performance on subsequent test frames.
There are two major challenges in the online adaptation framework. First, the lack of 3D annotations for test data increases the training difficulty at inference time. 
Second, we observe that the online adaptation can be easily influenced by the hard samples in the rapidly changing environment of the streaming test domain.

\begin{figure}[t]
    \centering
    \includegraphics[width=\linewidth]{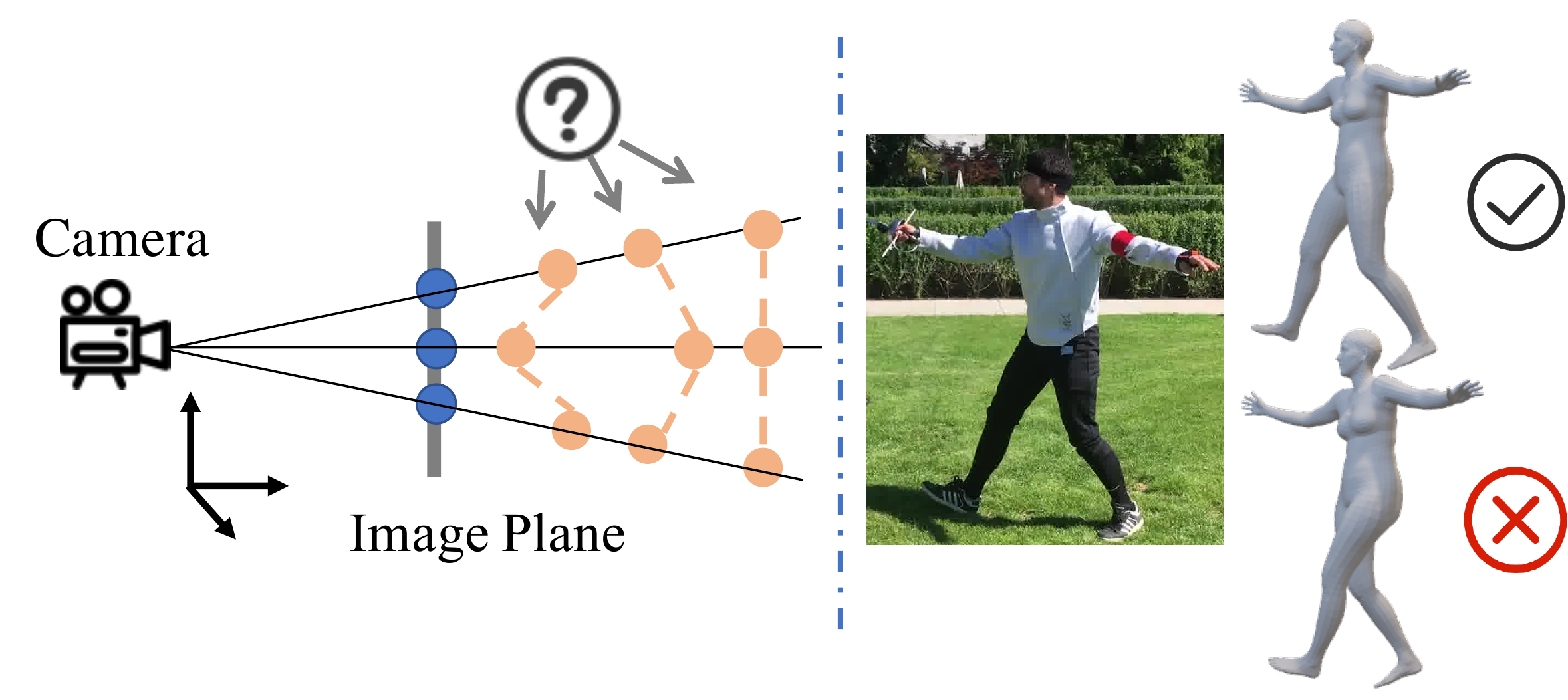}
    \caption{Adapting the model with 2D re-projection losses leads to 3D ambiguity, where multiple 3D vertices are projected to the same 2D position. It may result in wrong estimation of the posture.}
    \label{fig:motivation}
    \vspace{-5pt}
\end{figure}

For the first challenge, despite the lack of 3D annotations for test data, some optimization-based approaches~\cite{joo2020eft,bogo2016keep,SMPL-X:2019} provide alternative solutions that iteratively fit on each test frame through frame-wise losses, \eg, the pose re-projection loss of 2D keypoints\footnote{It is a common practice to use the ground-truth 2D keypoints for cross-domain human mesh reconstruction and pose estimation~\cite{joo2020eft,zhang2020inference,gong2021poseaug}}.
Notably, these objective functions are imperfect due to inevitable mismatches with 3D evaluation metrics, and therefore do not always lead to effective online adaptation.
For example, as shown in \fig{fig:motivation}, the optimization of 2D re-projection loss may lead to ambiguity in the learning process related to depth information, thus affecting the quality of mesh reconstruction.
To reduce the ambiguity, we first propose to use temporal constraints to regularize the training process of 2D losses.
However, we find that simply combining multiple objectives leads to undesirable results due to the competition and incompatibility between them, in the sense that the gradient of the loss terms may interfere with each other.
%
To solve this problem, we propose a new training approach named Bilevel Online Adaptation (\textbf{BOA}) that greatly benefits joint learning of frame-wise losses and temporal constraints.
Specifically, when a new frame is observed, the lower-level optimization step of BOA probes rational model parameters under frame-wise losses. On this basis, the upper-level optimization step is to find feasible responses to the overall objectives with temporal constraints and then update the model with approximated second-order derivatives \cite{finn2017model}.

To further reduce the mismatch between the online learning objective and the 3D evaluation metrics, in BOA, we leverage 3D annotations that are available and effective in the training set, and co-train the model with data samples from both source and target domains. As an extension of its predecessor at CVPR’2021~\cite{guan2021bilevel}, we efficiently perform feature retrieval on different clusters of the offline training set, and use the obtained source video frames with 2D postures similar to the target sample as training guidance.


For the second challenge, since the streaming video frames arrive asynchronously, the distribution of test data observed is usually non-stationary.
Therefore, it is difficult to strike a balance between avoiding over-fitting regular frames and avoiding under-fitting hard samples that are referred to as ``\textit{key-frames}'' and typically correspond to severe occlusion, sudden movement, and scene changes.
In this paper, we improve our previous work with a dynamic learning strategy termed as \textbf{DynaBOA}, which adaptively refines the upper-level optimization phase for key-frames through multiple training steps, so that the model can fully adapt to these hard samples.
The key idea is that when adapting the model on key-frames, the parameters would change dramatically responding to non-stationary data distribution, because the priors learned from regular frames cannot be fully reused.
Based on this fact, DynaBOA measures the distance between features before and after a round of bilevel optimization, and dynamically adjusts the number of training iterations on each frame until feature convergence.
%

This paper also makes a substantial extension to the experiments of its preliminary work.
We use Human 3.6M~\cite{h36m_pami} as the source domain and perform DynaBOA on three challenging human mesh reconstruction datasets with streaming video frames, namely 3DPW~\cite{vonMarcard2018}, MPI-INF-3DHP~\cite{mehta2017monocular}, and SURREAL~\cite{varol2017learning}. These datasets yield remarkable domain gaps from the source training set. 
Our approach significantly outperforms baselines in quantitative evaluation and qualitative results.
Further, we present analyses to validate the effectiveness of each component of DynaBOA.


\section{Related Work}

\subsection{3D Human Mesh Reconstruction}
SMPL~\cite{loper2015smpl} is a widely used parametric model for 3D human mesh reconstruction, which is also used in this work.
Early methods~\cite{guan2009estimating,sigal2008combined,bogo2016keep,lassner2017unite,huang2017towards,zanfir2018monocular} gradually fit a standard T-Pose SMPL model to an input image based on the silhouettes~\cite{lassner2017unite} or 2D keypoints~\cite{bogo2016keep}.
These optimization-based methods are time-consuming, struggling at the inference time on a single input.
Recently, many approaches~\cite{kanazawa2018end,Moon_2020_ECCV_I2L-MeshNet,kocabas2020vibe,kolotouros2019learning,aksan2019structured,xu2019denserac,guler2019holopose,pavlakos2018learning} use deep neural networks to regress the parameters of the SMPL model, which are efficient and can produce more accurate reconstruction results if large-scale 3D data is available.
However, most existing datasets with accurate 3D annotations are captured in constrained environments, such as a green screen studio. It's challenging for models trained on these datasets to generalize well to in-the-wild images.
%
To tackle this problem, Kanazawa~\etal~\cite{kanazawa2018end} proposed an adversarial training framework to utilize the unpaired 3D annotations to facilitate the reconstruction of int-the-wild data. 
Other approaches~\cite{sun2019human,tung2017self,li2019towards,omran2018neural,rueegg2020chained} improves the in-the-wild performance by designing more effective temporal features~\cite{sun2019human,tung2017self} or employing more informative input such as RGB-D~\cite{li2019towards} and segmentation images~\cite{omran2018neural,rueegg2020chained}.
Unlike the above methods, in this work, we further tackle this problem in out-of-domain streaming scenarios by using an online adaptation algorithm.

\subsection{Unsupervised Online Adaptation}
In this paper, we present a pilot study of unsupervised online adaptation in the context of human mesh reconstruction, which refers to sequentially adapting a pre-trained model to streaming test data without 3D labels of the target domain.
It is an emerging technique to prevent model crashing when the test data has a significant distribution shift from the training data.
Previous methods~\cite{duchi2011adaptive,broderick2013streaming,bobu2018adapting,liu2020learning,park2018meta,voigtlaender2017online,tonioni2019real,broderick2013streaming,tonioni2019learning,zhang2020online,li2020self} use the online adaptation framework for tasks other than mesh reconstruction, such as video segmentation~\cite{voigtlaender2017online}, tracking~\cite{park2018meta}, and stereo matching~\cite{tonioni2019real,tonioni2019learning}. 
Beyond unsupervised online adaptation, many previous approaches effectively learn generalizable features through meta-learning \cite{finn2017model,fallah2020convergence,Huang_2021_CVPR}, learning domain-invariant representations \cite{khosla2012undoing,muandet2013domain,ghifary2015domain,li2017deeper,li2018domain,li2019episodic}, or learning with adversarial examples \cite{shankar2018generalizing,volpi2018generalizing,yao2021videodg}.
However, none of these approaches have been focused on how to online adaptation on streaming data.
\section{Preliminaries}

\subsection{Problem Setup}
\label{sec:problem}

The SMPL-based solution to human mesh reconstruction can be usually specified as a tuple of $(\boldsymbol{X}, \boldsymbol{\Theta}, \boldsymbol{\Pi}_{\boldsymbol{C}}, \mathcal{L})$, where $\boldsymbol{X}$ denotes the observation space;
$\boldsymbol{\Theta}=\{\boldsymbol{\beta}, \boldsymbol{\theta}\}$ denotes the parameters of the SMPL model \cite{loper2015smpl}, where $\boldsymbol{\beta}$ and $\boldsymbol{\theta}$ correspond to the shape and the posture of human body respectively.
%
For each input frame $\boldsymbol{x}_t \in \boldsymbol{X}$, a first-stage model is trained to estimate $\boldsymbol{\Theta}_{t}$. 
Then the SMPL model generates the corresponding mesh and recovers 3D keypoints denoted by $\widehat{\boldsymbol{J}}_t$ using a mesh-to-3D-skeleton mapping pre-defined in SMPL.
%
The third element in the tuple is a weak-perspective projection model for projecting $\widehat{\boldsymbol{J}}_t$ to 2D space, \ie, $\hat{\boldsymbol{j}}_t = \boldsymbol{\Pi}_{\widehat{\boldsymbol{C}}_t}(\widehat{\boldsymbol{J}}_t)$, where $\widehat{\boldsymbol{C}}_t$ denotes the parameters of the weak-perspective projection estimated from $\boldsymbol{x}_t$.
The last one in the tuple defines a loss function $\mathcal{L}(\cdot)$ on $(\widehat{\boldsymbol{\beta}}_t, \widehat{\boldsymbol{\theta}}_t, \widehat{\boldsymbol{C}}_t,\widehat{\boldsymbol{J}}_t, \widehat{\boldsymbol{j}}_t)$ to learn the first-stage model $\mathcal{M}_{\boldsymbol{\phi}}$, usually in terms of neural networks.

In this work, we make two special modifications to the above task. First, we focus on out-of-domain scenarios, in the sense that large discrepancies may exist between the data distributions of the source training domain $\mathcal{D}^\text{src}$ and the target test domain $\mathcal{D}^\text{trg}$. 
Second, we specifically focus on dealing with streaming video frames $\{\boldsymbol{x}_t\}_{t=1}^{T}$ that arrive in a sequential order at test time.


\subsection{Online Adaptation Framework} 
\label{sec:online_adaptation}
This paper provides an early study of applying an online adaptation framework to test data to solve the out-of-domain 3D mesh reconstruction problem.
In this section, we introduce the general framework of test-time adaptation and provide a further discussion about its two main training difficulties.

We denote the pre-trained model from the source domain as $\mathcal{M}_{\boldsymbol{\phi}_0}$.
Note that we do not have special requirements for the pre-training method.
Typically, $\mathcal{M}_{\boldsymbol{\phi}_0}$ has an encoder to extract image features, and a regressor to estimate the parameters of SMPL and the camera configurations from the image features.

Given sequentially arrived target video frames $\{\boldsymbol{x}_t\}_{t=1}^T \in \mathcal{D}^\text{trg}$, a straightforward solution to quickly absorbing the domain-specific knowledge is to fine-tune $\mathcal{M}$ consecutively on each individual $\boldsymbol{x}_t$, following the online adaptation paradigm proposed by Tonioni~\etal~\cite{tonioni2019real}. 
We take it as a baseline algorithm that computes the loss function $\mathcal{L}$ with pose constraints\footnote{We discuss more about the specific forms of $\mathcal{L}$ in \sect{sec:boa}.} on each $\boldsymbol{x}_t$, and performs a single optimization step as follows before the inference step:
\begin{align}
    \boldsymbol{\phi}_{t} \leftarrow \boldsymbol{\phi}_{t-1} - \alpha \nabla_{\boldsymbol{\phi}} \mathcal{L}(\boldsymbol{x}_t; \mathcal{M}_{\boldsymbol{\phi}_{t-1}}),
    \label{eq:online_adaptation_baseline}
\end{align}
where $\alpha$ is the learning rate of gradient descent. There are two potential drawbacks in the baseline algorithm to be improved:
\begin{itemize}[leftmargin=10pt]
    \item Although fine-tuning a learned model on unlabeled target data may help to handle continuously changing test environments, due to \textbf{the lack of 3D supervisions}, an imperfect frame-wise loss may lead to wrong directions of gradient descent and thus increase the 3D ambiguities of the final results.
    \item Because of the non-stationarity of the streaming data, it is difficult to strike a balance between avoiding overfitting of regular frames, which will increase the training difficulty on subsequent videos, and avoiding \textbf{under-fitting of key-frames}, which correspond to severe occlusions, sudden movements, or scene changes. We observe that a temporally uniform optimization strategy, \ie, \eqn{eq:online_adaptation_baseline}, cannot effectively adapt the model to the hard cases in the key-frames, which may also have negative and long-term effects on the results of subsequent regular frames. 
\end{itemize}
\section{Approach}

\begin{figure*}[t]
    \centering
    \includegraphics[width=0.95\linewidth]{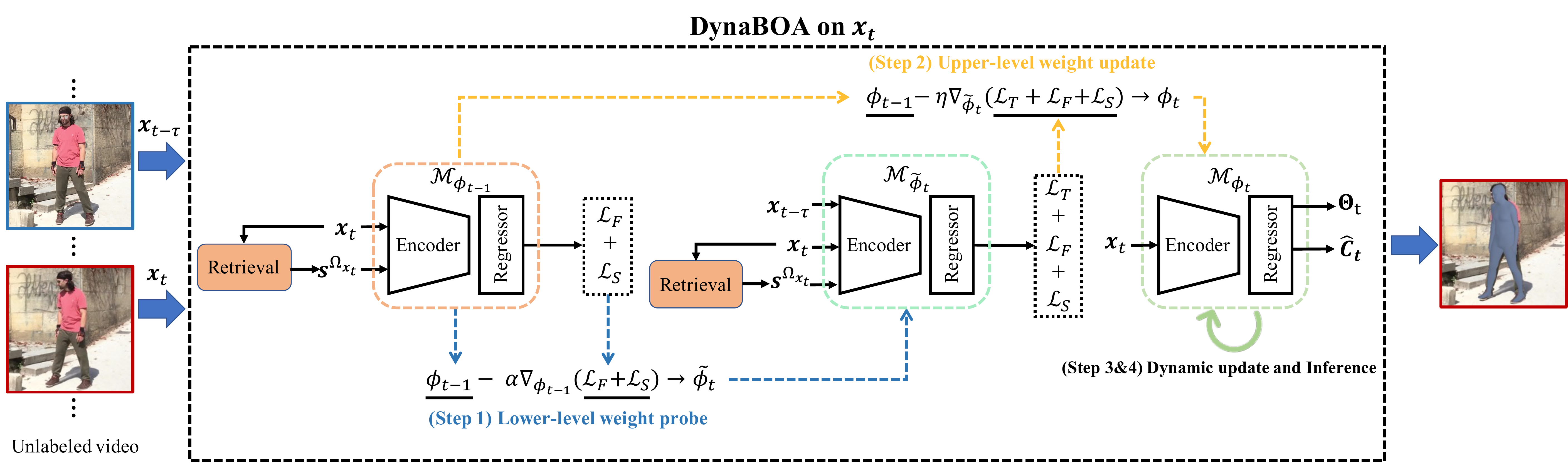}
    \caption{An overview of the proposed approach of Dynamic Bilevel Online Adaptation (DynaBOA), where the lower-level training step serves as a weight probe to find a feasible response to the frame-wise pose constraints, and the upper-level training step minimizes the overall multi-objectives in space-time and updates the model with approximated second-order derivatives. The model uses an adaptive number of optimization steps to adjust dynamically to address the non-stationary distribution shift over time. It also retrieves source exemplars as 3D guidance to compensate for the lack of target 3D supervision.
    Finally, the model updated by DynaBOA infers the parameters $\widehat{\boldsymbol{\Theta}}_{t}$ of SMPL and the camera configurations $\widehat{{\boldsymbol{C}}_{t}}$ of the current frame. 
   }
    \label{fig:framework}
\end{figure*}

In this section, we propose a new training approach named \textit{dynamic bilevel online adaptation} (DynaBOA) as a solution to out-of-domain human mesh reconstruction from streaming videos. \fig{fig:framework} gives an overview of the entire DynaBOA framework.
DynaBOA has the following three technical contributions to solve the aforementioned two challenges in online adaptation. For the first challenge of lacking 3D supervisions in the new domain:
\begin{itemize}[leftmargin=10pt]
    \item \textbf{Bilevel online adaptation} (\sect{sec:boa}): we introduce additional temporal constraints and alleviate the conflicts between the space-time multi-objectives through a bilevel online adaptation (BOA) procedure. 
    \item \textbf{3D exemplar guidance} (\sect{sec:exemplar_guidance}): It regularizes the training process on the streaming data by retrieving similar source exemplar, and including the well-defined 3D annotations in bilevel optimization. 
\end{itemize}
For the second challenge of under-fitting key-frames due to the non-stationarity of streaming data:
\begin{itemize}[leftmargin=10pt]
    \item \textbf{Dynamic update strategy} (\sect{sec:adaptive update}): We improve BOA with the dynamic update strategy, which fully adapts the model to hard streaming samples (termed as key-frames) while adaptively avoiding overfitting regular frames.
\end{itemize}
Notably, in both \sect{sec:exemplar_guidance} and \sect{sec:adaptive update}, DynaBOA makes substantial improvements over its predecessor at CVPR'21~\cite{guan2021bilevel}.

\begin{algorithm}[t] 
  \caption{Bilevel Online Adaptation}  
  \label{alg:BOA}  
  \small
  \begin{algorithmic}[1]
    \Require  
    Sequential frames $\{\boldsymbol{x}_t\}^{T}_{t=1}$ from the test set, a base model $\mathcal{M}_{\boldsymbol{\phi}_{0}}$ parameterized by $\boldsymbol{\phi}_{0}$, the learning rates $\alpha$ and $\eta$, the moving average rate $\gamma$ of the teacher model.
    \Ensure  
    SMPL and camera parameters $\widehat{\boldsymbol{\Theta}}_{t}$, $\widehat{\boldsymbol{C}}_{t}$.
    \State \textit{\# Initialize the teacher model}
    \State \textbf{Initialize} $\boldsymbol{\omega}_{0} \leftarrow \boldsymbol{\phi}_{0}$ 
    \For{$t=1, \ldots, T$}
        \State \textit{\# Lower-level weight probe}
        \State $\widetilde{\boldsymbol{\phi}}_{t} \leftarrow \boldsymbol{\phi}_{t-1} - \alpha \nabla_{\boldsymbol{\phi}}( \mathcal{L}_{F}(\boldsymbol{x}_{t};\boldsymbol{\phi}_{t-1}))$ 
            %
        \State \textit{\# Upper-level weight update}
        \State $\boldsymbol{\phi}_{t} \leftarrow \boldsymbol{\phi}_{t-1} - \eta \nabla_{\boldsymbol{\phi}} (\mathcal{L}_{T}(\boldsymbol{x}_{t}, \boldsymbol{x}_{t-\tau}; \boldsymbol{\omega}_{t-1}, \widetilde{\boldsymbol{\phi}}_{t})+\mathcal{L}_{F}(\boldsymbol{x}_t; \widetilde{\boldsymbol{\phi}}_{t}))$ 
        \State \textit{\# Update the teacher model}
        \State $\boldsymbol{\omega}_{t} \leftarrow \gamma \boldsymbol{\omega}_{t-1} + (1-\gamma) \boldsymbol{\phi}_{t}$
        \State \textit{\# Infer SMPL and camera parameters}
        \State $\widehat{\boldsymbol{\Theta}}_{t}$, $\widehat{\boldsymbol{C}}_t \leftarrow \mathcal{M}_{\boldsymbol{\phi}_{t}}(\boldsymbol{x}_{t})$
    \EndFor
  \end{algorithmic}  
\end{algorithm}

\subsection{Bilevel Optimization with Space-Time Constraints}
\label{sec:boa}


Bilevel Online Adaptation is the main body of the DynaBOA, which is illustrated in \alg{alg:BOA}.
%
In BOA, we introduce additional temporal constraints to compensate for the unavailable 3D annotations, and alleviate the conflicts between multiple training objectives through a bilevel optimization procedure.
%

\begin{figure}[t]
    \centering
    \includegraphics[width=0.95\linewidth]{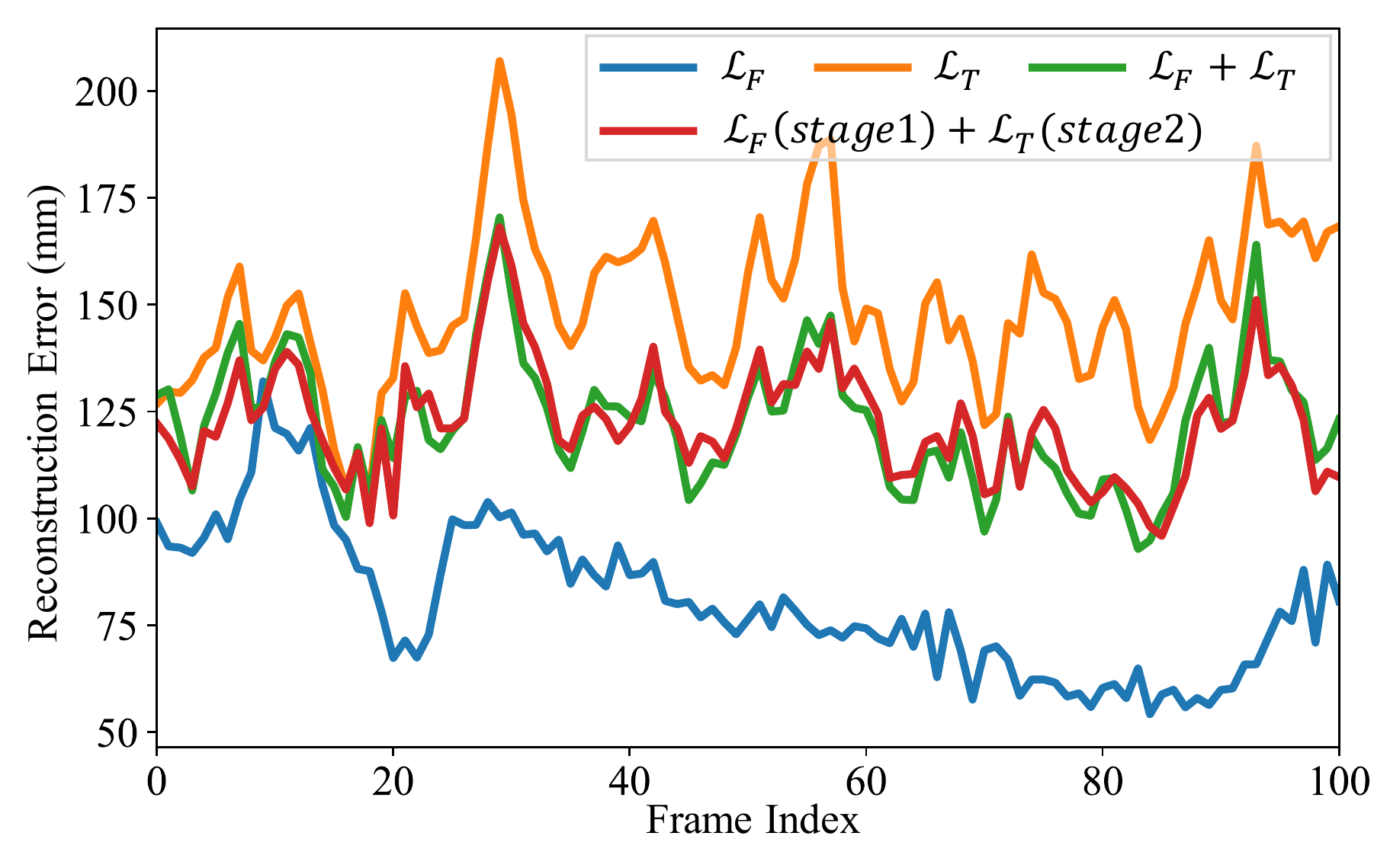}
    \vspace{-10pt}
    \caption{The reconstruction errors of various single-level optimization schemes, indicating the incompatibility between the space-time multi-objectives. 
    We use MPJPE as the evaluation metric.
    }
    \label{fig:loss_converge_compare}
\end{figure}

\myparagraph{Drawbacks of single-level optimization.} 
In our problem setup, there are naturally strong temporal dependencies between the streaming frames, which can be leveraged to improve the quality of out-of-domain online adaptation. Suppose that we have two objectives respectively for frame-wise shape and pose constraints ($\mathcal{L}_{F}$) and temporal consistency across frames ($\mathcal{L}_{T}$), whose specific forms are discussed later, straightforward methods to combine $\mathcal{L}_{F}$ and $\mathcal{L}_{T}$ include jointly optimizing them by adding them together or iteratively optimizing $\mathcal{L}_{F}$ and $\mathcal{L}_{T}$ in two stages.
However, as shown in \fig{fig:loss_converge_compare}, the above methods usually lead to sub-optimal results compared to the one only using $\mathcal{L}_{F}$. The possible reason is that there exists the competition and incompatibility between $\mathcal{L}_{F}$ and $\mathcal{L}_{T}$, in the sense that the gradient of the single-frame constraint may interfere with the training of the temporal one.
%
%
It motivates us to design the bilevel optimization method to prevent the overfitting of a single objective function and to make both the single-frame constraint and the time constraint work.

\subsubsection{Lower-Level Weight Probe}
We formulate the problem of identifying effective model weights under space-time multi-objectives as a bilevel optimization problem. 
In this setup, as shown in \fig{fig:framework}, the lower-level optimization step serves as a weight probe to rational models under single-frame pose constraints, while the upper-level optimization step finds a feasible response to temporal constraints.
Specifically, for the $t$-th test sample, the model from the last online adaptation step, denoted by $\mathcal{M}_{\boldsymbol{\phi}_{t-1}}$, is firstly optimized with the single-frame constraints, $\mathcal{L}_{F}$, to obtain a set of temporary weights denoted by $\widetilde{\phi}_t$. 
We name this procedure as the \textit{lower-level probe} (\underline{Line 5} in \alg{alg:BOA}), in the sense that first, $\widetilde{\phi}_t$ can be feasible responses to the easy component of multi-objectives with respect to pose priors, which best facilitates the rest of the learning procedure for temporal consistency; Second, $\widetilde{\phi}_t$ is not directly used to update $\mathcal{M}_{\boldsymbol{\phi}_{t-1}}$. 
At this level, we focus on the spatial constraints on individual frames:
\begin{align}
    \mathcal{L}_{F} = \mu_1 ||\boldsymbol{j}_t - \widehat{\boldsymbol{j}_t}||_2^2 + \mu_2 \rho(\widehat{\boldsymbol{\beta}}_t, \widehat{\boldsymbol{\theta}}_t),
    \label{Eq:streaming_adapt}
\end{align}
where $\{\mu_1, \mu_2\}$ are the loss weights. The first loss term is the supervision of the re-projection error of 2D keypoints. The second term is the prior constraint on the shape and pose parameters $\boldsymbol{\Theta}=\{\boldsymbol{\beta}, \boldsymbol{\theta}\}$, which is a common practice in 3D human mesh reconstruction. Here, $\rho(\cdot)$ calculates the distance of the estimated $\widehat{\boldsymbol{\beta}}_t, \widehat{\boldsymbol{\theta}}_t$ to their statistic priors\footnote{These priors are obtained from a commonly-used third-party database.}. 
The lower-level optimization step produces a \textit{hypothetic} model $\mathcal{M}_{\widetilde{\phi}_t}$ for subsequent upper-level optimization step but does not update $\mathcal{M}_{\boldsymbol{\phi}_{t-1}}$. 
Note that, due to the lack of 3D supervisions in the target domain, the above $\mathcal{L}_{F}$ is insufficient to recover the 3D body (see \fig{fig:motivation}) where multiple 3D vertices may be projected to the same 2D position. 
Therefore, it is a potentially effective method to reduce 3D ambiguity by using the multi-frame information in the streaming data and optimizing the time consistency of the reconstruction results.

\subsubsection{Upper-Level Weight Update}
At the upper-level BOA step, we calculate the overall space-time multi-objectives using $\mathcal{M}_{\widetilde{\phi}_t}$ obtained from lower-level optimization, and then perform backpropagation to update the original model $\boldsymbol{\phi}_{t-1}$ (\underline{Line 7} in \alg{alg:BOA}).
As for the specific form of the temporal constraints, given two images $(\boldsymbol{x}_{t-\tau}, \boldsymbol{x}_{t})$ at a time interval $\tau$, we define a motion loss on their 2D keypoints $(\boldsymbol{j}_{t-\tau}, \boldsymbol{j}_t)$ and the estimation $(\widehat{\boldsymbol{j}}_{t-\tau}, \widehat{\boldsymbol{j}}_t)$ by $\mathcal{M}_{\widetilde{\phi}_t}$:
\begin{align}
    \mathcal{L}_\text{motion} &= \big\|\big(\widehat{\boldsymbol{j}}_t - \widehat{\boldsymbol{j}}_{t-\tau}\big) - \big(\boldsymbol{j}_t - \boldsymbol{j}_{t-\tau}\big)\big\|_2^2.
\end{align}
Another part of the temporal constraints is to overcome the effect of \textit{catastrophic forgetting} of target distribution and to avoid overfitting the current frames. We thus maintain a teacher model $\mathcal{T}_\omega$ parametrized by the exponential moving average of historical $\phi_t$ (\underline{Line 9} in \alg{alg:BOA}).
Inspired by the work of \textit{Mean Teacher}~\cite{tarvainen2017mean}, we regularize the output of $\mathcal{M}_{\widetilde{\phi}_t}$ to be consistent with $\mathcal{T}_{\boldsymbol{\omega}_{t-1}}$: 
\begin{align}
    \mathcal{L}_\text{teacher} = \big\|\mathcal{T}_{\boldsymbol{\omega}_{t-1}}(\boldsymbol{x}_t) - \mathcal{M}_{\widetilde{\phi}_t}(\boldsymbol{x}_t)\big\|_2^2.
\end{align}
$\mathcal{L}_\text{teacher}$ is then combined with the motion loss to obtain the overall temporal constraints:
\begin{align}
    \mathcal{L}_{T} = \mu_3 \mathcal{L}_{\text{motion}} + \mu_4 \mathcal{L}_{\text{teacher}},
    \label{Eq:t_loss}
\end{align}
where $\mu_3$ and $\mu_4$ indicate the weights of the temporal loss terms. 
These two losses are complementary: the regularization from the teacher model maintains long-term temporal information, while the motion loss focuses more on short-term consistency. 
As shown in \alg{alg:BOA}), we finally update the model parameter based on $\boldsymbol{\phi}_{t-1}$ with both the frame-wise losses in \eqn{Eq:streaming_adapt} and the temporal losses in \eqn{Eq:t_loss}. 
By these means, BOA avoids overfitting the temporal constraints by retaining the pose prior loss for the upper optimization level. 
On the other hand, it avoids overfitting the pose priors by updating the model weights based on $\boldsymbol{\phi}_{t-1}$ instead of $\widetilde{\phi}_t$.
As a result, BOA effectively combines single-frame and temporal constraints, achieving considerable improvement over the straightforward solutions mentioned above.

\subsection{3D Exemplar Guidance}
\label{sec:exemplar_guidance}


In addition to using temporal constraints to effectively reduce depth ambiguity, in this part, we consider another solution from a new perspective to compensate for the lack of 3D annotations in the target domain.
The basic idea is to introduce complete 3D supervisions into the BOA training process by performing cross-domain feature retrieval for source exemplars, and then train the model jointly on streaming test data and the retrieved source data. 
Despite the notable distribution shift between the source and target domains, we argue that some source samples, which contain similar shapes and postures to the streaming data, can be used to ease the online adaptation process. They have well-defined 3D annotations that can provide accurate 2D-to-3D exemplar guidance. By leveraging them in BOA, we can greatly reduce the impact of inconsistencies between the objective functions of model training and the measurements of model evaluation.

\begin{figure}[t]
    \centering
    \includegraphics[width=\linewidth]{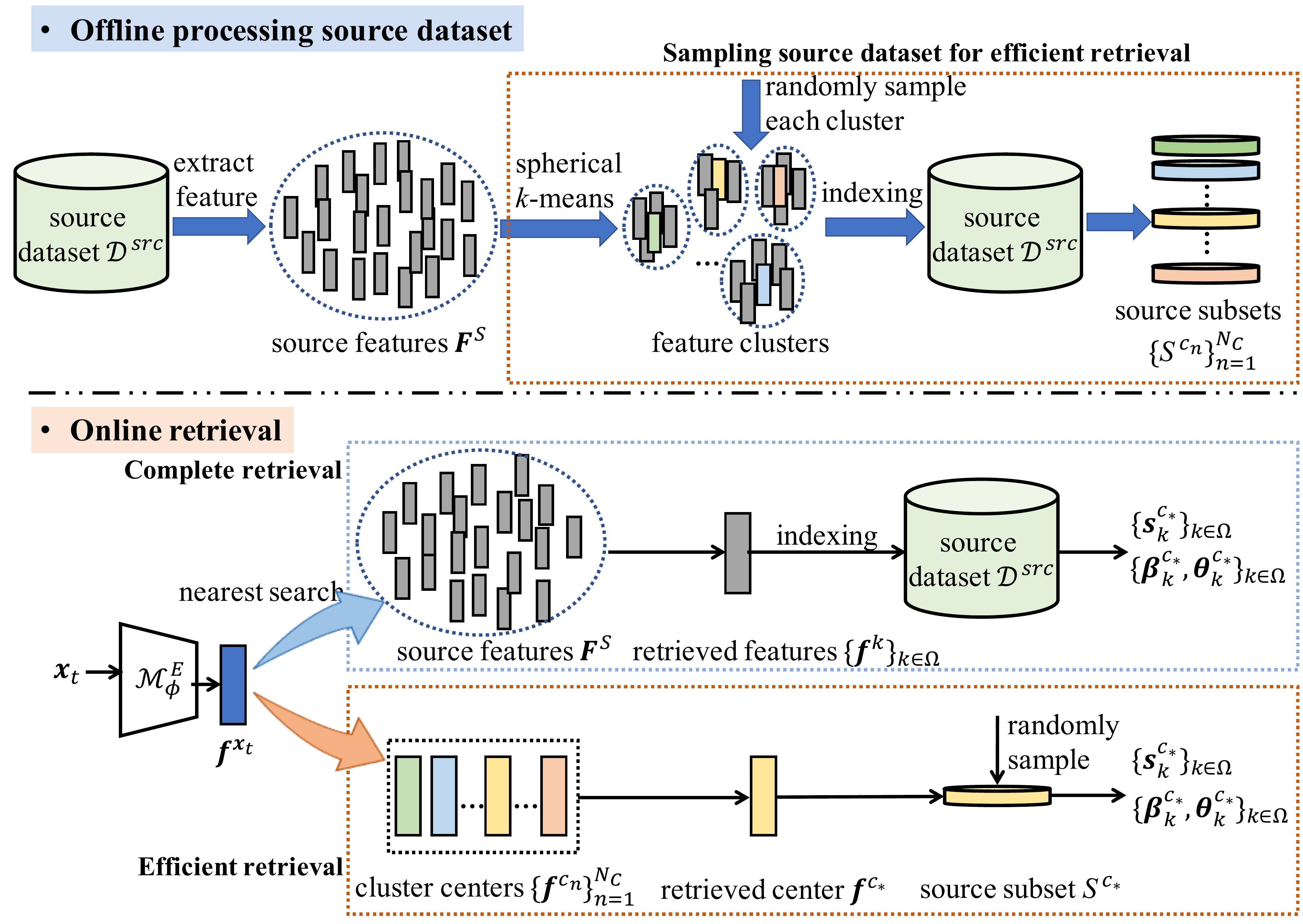}
    \vspace{-15pt}
    \caption{
    Illustration of the cross-domain retrieval for 3D example guidance. \textbf{Top:} The source data is processed offline. \textbf{Bottom:} A comparison of the efficiency between the baseline method of complete retrieval and the proposed retrieval method. ``indexing'' refers to finding the data with the same index in the source domain.}
    \label{fig:method_retrieval}
\end{figure}

\begin{figure}[t]
    \centering
    \includegraphics[width=\linewidth]{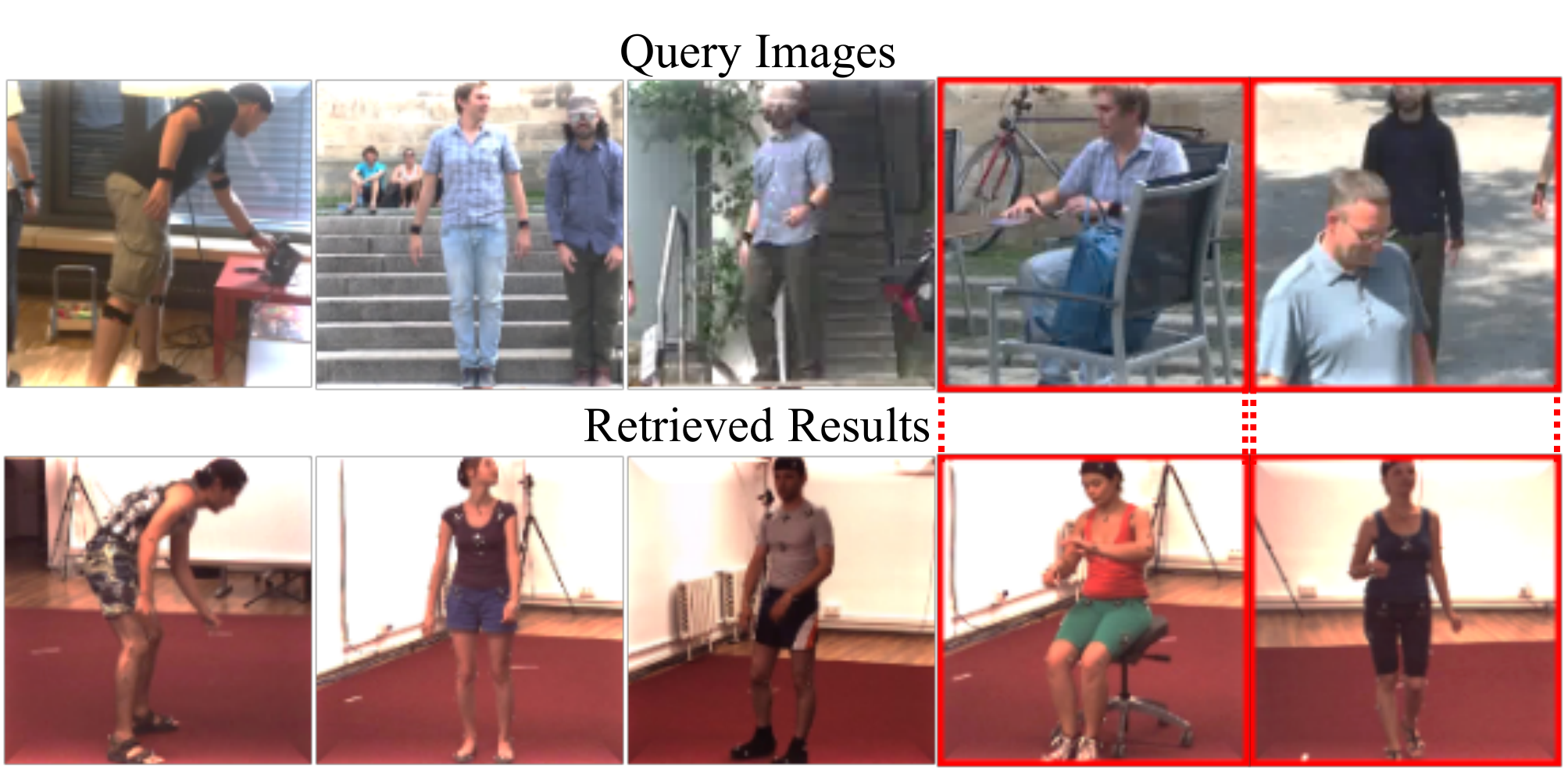}
    \vspace{-15pt}
    \caption{Showcases of the query images from the target domain and the corresponding retrieved images from the source domain. The retrieved results have similar postures to the query images, even though the persons in the query images are severely occluded (highlighted in the red boxes in the two columns on the right).
    }
    \label{fig:retrieval_vis}
\end{figure}

In the conference paper~\cite{guan2021bilevel}, we randomly select the source data from the whole source set. Although the use of random selection gains improvement, it inevitably brings in negative samples that are very different from the current target sample in terms of posture, camera viewpoint, and~\etc, which may partly reduce the ability of BOA in mitigating the distribution shift.
Therefore, we improve the random selection with a retrieval method that is based on the similarity between samples across domains. 
\fig{fig:method_retrieval} shows the overview of the retrieval procedure.
We conduct retrieval in the latent space of $\mathcal{M}_{\boldsymbol{\phi}}$ to avoid interference from the noises in the observation space, \eg, lighting and backgrounds.
In \fig{fig:retrieval_vis}, we visualize the retrieved results to analyze the effect of exemplar retrieval. 
By comparing the query images in the first row and retrieved images in the second row, we can observe that performing cross-domain retrieval in the latent space of $\mathcal{M}_{\boldsymbol{\phi}}$ obtains results with similar postures, even when the query image contains a severely occluded person or a noisy background. 

The entire retrieval method includes two processing stages that are respectively conducted offline and online. 
In the offline processing stage, we first use $\mathcal{M}^{\text{E}}_{{\phi}_0}$, which is the encoder of the source model $\mathcal{M}_{\boldsymbol{\phi}_0}$, to extract the source features $\boldsymbol{F}^S= \{\boldsymbol{f}^{\boldsymbol{s}_{j}}\}_{j=1}^{N_S}$ from the source dataset $\mathcal D^\text{src}$. 
$N_S$ is the number of source data.
$\boldsymbol{f}^{\boldsymbol{s}_j} \in \mathbb{R}^{d_f}$ is the feature of the $j$-th image $\boldsymbol{s}_j \in \mathcal D^\text{src}$.
As shown in the middle blue box in \fig{fig:method_retrieval}, a na\"{i}ve method is to directly search in the entire set of offline source features, termed as \textit{complete retrieval}.
However, since the source set is generally large (\eg, the Human3.6M~\cite{h36m_pami} dataset used in our work contains about $3.5$ million samples), the complete retrieval strategy requires massive memory resources and time costs that are not feasible in online adaptation.
Therefore, in the offline processing stage, we further group the source features $\boldsymbol{F}^{S}$ into $N_C$ clusters by spherical \textit{k}-means (see the upper orange box in \fig{fig:method_retrieval}).
The cluster centers are denoted as $\{\boldsymbol{f}^{\boldsymbol{c}_n}\}_{n=1}^{N_C}$.
We then randomly sample a small number of features from each cluster and use corresponding source images to form $N_C$ subsets $\{\mathcal{S}^{\boldsymbol{c}_{n}}\}_{n=1}^{N_C}$.

In the online retrieval stage, as shown in the bottom orange box in \fig{fig:method_retrieval}, the feature extracted by $\mathcal{M}^{\text{E}}_{{\phi}}$ from a query image $\boldsymbol{x}_{t} \in \mathcal{D}^{\text{trg}}$ is denoted as $\boldsymbol{f}^{\boldsymbol{x}_t}$. 
Here, $\mathcal{M}^{\text{E}}_{{\phi}}$ is the encoder of the entire online adaptation model parameterized by either $\mathcal{M}_{\phi_{t-1}}$ or $\mathcal{M}_{\widetilde{\phi}_{t}}$, since the retrieval process is performed in both optimization steps of \textit{lower-level weight probe} and \textit{upper-level weight update} in DynaBOA. 
%
Next, we search the nearest cluster $\mathcal{S}^{\boldsymbol{c}_{\ast}}$ by calculating the cosine similarity between $\boldsymbol{f}^{\boldsymbol{x}_t}$ and the features at all cluster centers. 
\begin{equation}
\begin{split}
    \mathcal{S}^{\boldsymbol{c}_{\ast}} = \mathop{\arg\max}\limits_{n} \ \frac{\boldsymbol{f}^{\boldsymbol{x}_t} \cdot \boldsymbol{f}^{\boldsymbol{c}_n}}{\big\|\boldsymbol{f}^{\boldsymbol{x}_t}\big\|\, \big\|\boldsymbol{f}^{\boldsymbol{c}_n}\big\|},
\end{split}
\end{equation}
where $\|\cdot\|$ is $\ell_2$ norm. In the following sections, we use $\texttt{Sim}(\cdot)$ to represent the cosine similarity.
Next, we randomly sample $K$ images from $\mathcal{S}^{\boldsymbol{c}_{\ast}}$ as the auxiliary training data, which are denoted as $\{{\boldsymbol{s}^{\boldsymbol{c}_{\ast}}_k}\}_{k \in \Omega}$. Here, $\Omega$ denotes the collection of indexes corresponding to the $K$ selected source images.
The efficient retrieval scheme has two advantages over the complete retrieval strategy that searches the entire source dataset (see \tab{tab:retrieval_compar}): First, it largely reduces time costs since we only calculate the similarity $N_C$ ($\ll N_S$) times between the streaming data and the cluster centers; Second, the proposed retrieval scheme is more memory-efficient, since we only store smaller subsets instead of the entire source training set.

Given the retrieved source images $\{{\boldsymbol{s}^{\boldsymbol{c}_{\ast}}_k}\}_{k \in \Omega}$ and the corresponding 3D SMPL annotations $\{{\boldsymbol{\beta}^{\boldsymbol{c}_{\ast}}_k}, {\boldsymbol{\theta}^{\boldsymbol{c}_{\ast}}_k}\}_{k \in \Omega}$, we train the model $\mathcal{M}_{\phi}$ using the following loss function: 
\begin{equation}
\small
    \begin{split}
    \mathcal{L}_S=\lambda_1\big\|\widehat{\boldsymbol{\beta}} - \boldsymbol{\beta}\big\|_2^2 + \lambda_2\big\|\widehat{\boldsymbol{\theta}} - \boldsymbol{\theta}\big\|_2^2 +\lambda_3\big\|\widehat{\boldsymbol{J}} - \boldsymbol{J}\big\|_2^2 + \lambda_4\big\|\widehat{\boldsymbol{j}} - \boldsymbol{j}\big\|_2^2,    
    \end{split}
    \label{eq:retri_loss}
\end{equation}
where we omit the superscripts and subscripts for simplicity. 
$\boldsymbol{J}$ denotes the 3D keypoints generated from $\{\boldsymbol{\beta}, \boldsymbol{\theta}\}$ by a mapping approach pre-defined in SMPL, and $\boldsymbol{j}$ indicates the ground-truth 2D keypoints. 
In DynaBOA, we co-train the model using \eqn{eq:retri_loss} upon the retrieved source data and using the proposed space-time constraints upon the streaming frames.
With \eqn{eq:retri_loss}, the model can obtain better 3D guidance to avoid depth ambiguity, and make the overall objective function better match 3D evaluation metrics.

\begin{table}[t]
    \centering
    \caption{Comparison of different retrieval strategies. We here use Human3.6M as an example source dataset.}
    \vspace{-5pt}
    \resizebox{\linewidth}{!}{
    \begin{tabular}{llccc}
    \toprule
    Strategy & Search space   & \# Compare & Time (ms/step) &Memory \\ 
    \midrule
    Complete & All source data &$N_S$ ($312$ k) &$1.3\times 10^{3}$ & $2.47$ GB\\ 
    Efficient & Cluster centers &$N_C$ ($0.1$ k)  &$4.3\times 10^{-1}$       &$0.24$ GB\\ 
    \bottomrule
    \end{tabular}
    }
    \label{tab:retrieval_compar}
\end{table}

\begin{figure*}
    \centering
    \includegraphics[width=\linewidth]{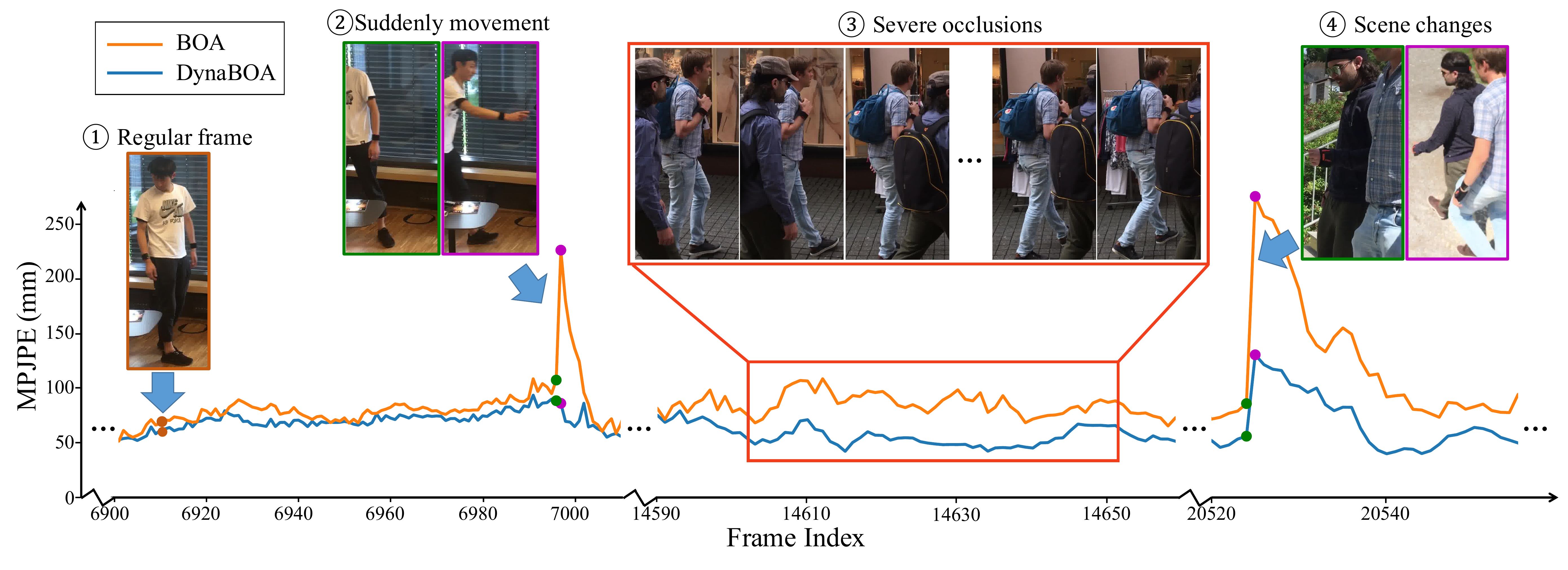}
    \vspace{-20pt}
    \caption{
    Performance of BOA (\alg{alg:BOA}) and DynaBOA (\alg{alg:adaptive_update}) on non-stationary streaming data, where the \textit{key-frames} correspond to the hard cases that can drastically increase the error of human mesh reconstruction, \eg, sudden movements, severe occlusions, and scene changes. Thanks to the dynamic update strategy, DynaBOA fully adapts to the hard samples while avoiding overfitting regular frames adaptively.
    }
    \label{fig:hard_case}
\end{figure*}

\begin{algorithm}[t] 
  \caption{Dynamic update strategy}  
  \label{alg:adaptive_update}  
  \small
  \begin{algorithmic}[1]
    \Require The $t$-th target frame $\boldsymbol{x}_t$, the models $\mathcal{M}_{\boldsymbol{\phi}_{t-1}}$ and $\mathcal{M}_{\boldsymbol{\phi}_{t}}$ before and after BOA (see \alg{alg:BOA}), the teacher model $\mathcal T_{\boldsymbol{\omega}_{t}}$ with a moving average rate $\gamma$, the feature distance threshold $\varepsilon$, the maximum steps of dynamic update $L$, the learning rate $\eta$.
    \Ensure $\mathcal{M}_{\boldsymbol{\phi}_t}$
    \State \textit{\# Extract features before and after BOA}
    \State $\boldsymbol{f}_{\boldsymbol{\phi}_{t-1}}^{\boldsymbol{x}_t}, \boldsymbol{f}_{\boldsymbol{\phi}_{t}}^{\boldsymbol{x}_t} \gets \mathcal{M}^\text{E}_{\boldsymbol{\phi}_{t-1}}(\boldsymbol{x}_t), \mathcal{M}^\text{E}_{\boldsymbol{\phi}_{t}}(\boldsymbol{x}_t)$
    \State \textit{\# Calculate the feature distance}
    \State $d_t \gets 1 - \texttt{Sim}(\boldsymbol{f}_{\boldsymbol{\phi}_{t}}^{\boldsymbol{x}_t}, \boldsymbol{f}_{\boldsymbol{\phi}_{t-1}}^{\boldsymbol{x}_t})$
    \State $l \gets 0$
    \While{$d_t > \varepsilon$ and $l < L$}
        \State \textit{\# For 3D exemplar guidance, refer to \sect{sec:exemplar_guidance}}
        \State $\{{\boldsymbol{s}^{\boldsymbol{c}_{\ast}}_k}\}_{k \in \Omega} \leftarrow \texttt{Retrieval}(\boldsymbol{x}_t, \{{\boldsymbol{s}^{\boldsymbol{c}_{\ast}}_k}\}_{k \in \Omega}; \boldsymbol{\phi}_{t})$
        \State \textit{\# Model refinement with upper-level objectives}
        \State $\boldsymbol{\phi}_{t}^{\prime} \leftarrow \boldsymbol{\phi}_{t} - \eta \nabla_{\boldsymbol{\phi}} (\mathcal{L}_{T}(\boldsymbol{x}_t, \boldsymbol{x}_{t-\tau}; \boldsymbol{\omega}_{t-1}, \boldsymbol{\phi}_{t}) + \mathcal{L}_{F}(\boldsymbol{x}_t; \boldsymbol{\phi}_{t}) + \mathcal{L}_{S}(\{{\boldsymbol{s}^{\boldsymbol{c}_{\ast}}_k}\}_{k \in \Omega};\boldsymbol{\phi}_{t})))$
        \State \textit{\# Re-calculate the feature distance}
        \State $\boldsymbol{f}_{\boldsymbol{\phi}^{\prime}_{t}}^{\boldsymbol{x}_t} \gets \mathcal{M}^\text{E}_{\boldsymbol{\phi}^{\prime}_{t}}(\boldsymbol{x}_t), \ d_t \gets 1 - \texttt{Sim}(\boldsymbol{f}_{\boldsymbol{\phi}_{t}}^{\boldsymbol{x}_t}, \boldsymbol{f}_{\boldsymbol{\phi}^{\prime}_{t}}^{\boldsymbol{x}_t})$
        \State $\boldsymbol{f}_{\boldsymbol{\phi}_{t}}^{\boldsymbol{x}_t} \gets \boldsymbol{f}_{\boldsymbol{\phi}^{\prime}_{t}}^{\boldsymbol{x}_t}, \ {\phi}_t \gets \boldsymbol{\phi}_t^{\prime}$
        \State \textit{\# Update the teacher model}
        \State $\boldsymbol{\omega}_{t} \leftarrow \gamma \boldsymbol{\omega}_{t} + (1-\gamma) \boldsymbol{\phi}_{t}$
        \State $l \gets l+1$
    \EndWhile
  \end{algorithmic}  
\end{algorithm}

\subsection{Dynamic Update Strategy}
\label{sec:adaptive update}

To cope with the \textit{key-frames} challenge mentioned in \sect{sec:online_adaptation}, we further improve the adaptation ability of BOA with a dynamic update strategy. 
Note that the key-frames correspond to the hard cases for human mesh reconstruction in non-stationary streaming data, \eg, sudden movements, severe occlusions, and scene changes. 
From the orange curve in \fig{fig:hard_case}, we can see that the reconstruction error of the original BOA~\cite{guan2021bilevel} drastically increases at key-frames, indicating a severe under-fitting problem.

Intuitively, since the distribution of the observed streaming data is highly non-stationary, in the sense that the hard cases in the key-frames are beyond what has been previously learned, the model may need more than one BOA step to explore and handle the significant distribution shift at key-frames.
A straightforward solution is to provide BOA with uniformly more training iterations at every frame, which, however, would greatly affect the computational efficiency online or might make the model unnecessarily overfit the regular frames with clear backgrounds, postures, and movements. 
Therefore, we improve BOA with a dynamic update strategy, which dynamically and adaptively adjusts the number of optimization steps according to the difficulty of each streaming sample.

\alg{alg:adaptive_update} illustrates the procedure of the dynamic update strategy.
%
The core of the dynamic update strategy is to measure the feature distance before and after a round of bilevel adaptation to determine whether the model $\mathcal{M}_{\boldsymbol{\phi}_{t}}$ needs to be refined on the current image (\underline{Lines 2-6} in \alg{alg:adaptive_update}).
Specifically, given frame $\boldsymbol{x}_t$, we first extract its feature $\boldsymbol{f}_{\boldsymbol{\phi}_{t-1}}^{\boldsymbol{x}_{t}}$ before bilevel adaptation by $\mathcal{M}^\text{E}_{\boldsymbol{\phi}_{t-1}}$, which is the encoder of the model at time step $t-1$.
After bilevel optimization, we can use the encoder $\mathcal{M}^\text{E}_{\boldsymbol{\phi}_t}$ of the adapted model to extract $\boldsymbol{f}_{\boldsymbol{\phi}_t}^{\boldsymbol{x}_{t}}$. 
Then, we calculate the cosine distance between $\boldsymbol{f}_{\boldsymbol{\phi}_{t-1}}^{\boldsymbol{x}_{t}}$ and $\boldsymbol{f}_{\boldsymbol{\phi}_t}^{\boldsymbol{x}_{t}}$ as $d_t = 1 - \texttt{Sim}(\boldsymbol{f}_{\boldsymbol{\phi}_{t}}^{\boldsymbol{x}_t}, \boldsymbol{f}_{\boldsymbol{\phi}_{t-1}}^{\boldsymbol{x}_t})$.
If $d_t$ is larger than a predefined threshold $\varepsilon$, $\mathcal{M}_{\boldsymbol{\phi}_{t}}$ will be refined on $\boldsymbol{x}_t$ with the same objectives used at the upper-level update step.
We repeat the refinement step on $\boldsymbol{x}_t$ (\underline{Lines} 8-10 in \alg{alg:adaptive_update}) and calculate the feature distance (\underline{Line} 12 in \alg{alg:adaptive_update}) until it is smaller than $\varepsilon$.
The main idea is that the model should change greatly if the input is a key-frame since it violates the previously learned test distribution, but for a regular frame, the model should change only slightly from the last time step.
Moreover, to avoid excessive time costs, we set the maximum number of model refinement on each frame to $L$.
As shown in \fig{fig:hard_case}, for key-frames, DynaBOA can produce more stable and accurate reconstruction results than its predecessor. Meanwhile, DynaBOA avoids overfitting the regular frames, which may cause training difficulties on subsequent streaming data.

\section{Experiments}
\label{sec:expri}

\begin{table*}[t]
    \centering
    \caption{Typical domain gaps between the source dataset (Human3.6M) and the target datasets used in our experiments in terms of focal length, bone length, camera distance, and camera height~\cite{DBLP:journals/corr/abs-2004-03143}.}
    \vspace{-5pt}
    \begin{tabular}{lllccccccc}
    \toprule
         Dataset        &\makecell[l]{Type}     &\makecell[l]{Capture device/software}   &\makecell[c]{\#Subjects}  &\makecell[c]{\#Frames} &\makecell[c]{Bone \\ len. (m)} &\makecell[c]{Focal \\ len. (pixel)}    &\makecell[c]{Camera \\dist. (m)}   &\makecell[c]{Camera\\ ht. (m)}\\ 
    \midrule
         Human3.6M    &Indoor                  &VICON~\cite{vicon}                          &5                        &312188  &3.9$\pm$0.1                    &1146.8$\pm$2.0                         &5.2$\pm$0.8                        &1.6$\pm$0.1\\
         \midrule
         3DPW          &In-the-wild             &Moving camera \& IMUs        &5                        & See \tab{tab:setting_diff} &3.7$\pm$0.1                    &1962.2$\pm$1.5                         &3.5$\pm$0.7                        &0.6$\pm$0.8\\
         MPI-INF-3DHP  &Indoor \& In-the-wild   &The Captury~\cite{captury}                    &6                        &2929  &3.7$\pm$0.1                    &1497.9$\pm$2.8                         &3.8$\pm$0.8                        &0.8$\pm$0.4\\
         SURREAL       &Synthetic               &Blender~\cite{blender}                        &30                       &19500  &3.7$\pm$0.2                   &600 $\pm$ 0                            &8.0$\pm$1.0                        &0.9$\pm$0.1\\
    \bottomrule
    \end{tabular}
    \label{tab:dataset_gap}
\end{table*}

\subsection{Implementation Details} \label{sec:dataset}

We provide the source code and video results at \url{https://sites.google.com/view/dynaboa}, and include descriptions of network architectures and training details in the \underline{Supplementary Materials}.

\myparagraph{Datasets.} 
We train the source model $\mathcal{M}_{\phi_0}$ using the Human3.6M dataset~\cite{h36m_pami} and respectively adapt the model to three test sets: 3DPW~\cite{vonMarcard2018}, MPI-INF-3DHP~\cite{mehta2017monocular}, and SURREAL~\cite{varol2017learning}. Table~\ref{tab:dataset_gap} presents the statistics of typical domain gaps among these datasets.
\begin{itemize}[leftmargin=10pt]
    \item \textbf{Human3.6M}~\cite{h36m_pami} is captured in a red-screen studio with $4$ digital video cameras. It has $11$ subjects in total. Like existing approaches~\cite{kocabas2020vibe,kanazawa2018end}, we train the base model on $5$ subjects (S1, S5--S8) and down-sample all videos from $50$fps to $10$fps.
    \item \textbf{3DPW}~\cite{vonMarcard2018} is a large-scale in-the-wild dataset captured by a moving phone camera and IMU sensors. Different from Human3.6M, 3DPW contains dynamic scenes with serve occlusions by objects or humans. Also, it includes novel actions beyond the scope of Human3.6M, such as climbing and fencing. We use its test set as the streaming target domain. 
    \item \textbf{MPI-INF-3DHP (3DHP)}~\cite{mehta2017monocular} is captured with wide field-of-view cameras, leading to strong camera distortion. Like above, we use its test set as the streaming target domain, which has $2{,}929$ images. It consists of $3$ dynamic scenes, \ie, in the studio with or without a green screen, and outdoor, each with $2$ subjects. All of the following factors increase the data non-stationarity of this benchmark and expand its distribution shift from Human3.6M, including various camera configurations, more complex environments, and novel actions.
    \item \textbf{SURREAL}~\cite{varol2017learning} is a large-scale synthetic dataset with ground-truth SMPL annotations, where most video clips have $100$ frames. In addition to the appearance gap between the synthetic images and the real images in Human3.6M, SURREAL is more diverse in body shapes and postures, backgrounds, and camera views. Unlike the previous literature~\cite{varol2018bodynet,zeng20203d} that typically uses the middle frames of the validation sequences to construct the test set, we take the full $100$-frame sequences as the streaming test set that has $19{,}500$ frames in total. All compared models are evaluated on the same test split.
\end{itemize}

\myparagraph{Evaluation details.}
We use multiple evaluation metrics on the target datasets, including MPJPE, PA-MPJPE, PVE, PCK, and AUC. More details can be found in the \underline{Supplementary Materials}. 
Particularly on the 3DHP dataset, we follow the existing approaches~\cite{kolotouros2019learning, kanazawa2018end} to evaluate the reconstruction results at the $17$ body joints.
On the 3DPW dataset, we adopt the data preprocessing protocols (\ie, \#PH, \#PV, and \#PS) from HMMR~\cite{DBLP:conf/cvpr/KanazawaZFM19}, VIBE~\cite{kocabas2020vibe}, and SPIN~\cite{kolotouros2019learning}.
The protocols differ in two aspects (see \tab{tab:setting_diff}). First, unlike \#PS and \#PV that adopt the original SMPL labels, \#PH uses the fitted neutral meshes as the ground truth, which are generated by minimizing the Euclidean vertex error between the fitted meshes and the original meshes. Second, the protocols have different criteria to select valid frames, resulting in different numbers of test frames.
We find that the use of different preprocessing protocols has a significant impact on the final results (see \tab{tab:3dpw}). Therefore, we present results under all three protocols and compare them with those of the state-of-the-art methods under the corresponding protocols.

\begin{table}[t]
    \centering
    \caption{Illustration of different protocols for data preprocessing on the 3DPW dataset, which have great impacts on the final resuls. \#PV/\#PS use the SMPL annotations from the original 3DPW, while \#PH uses the fitted neutral labels excluding all gender information.}
    \vspace{-5pt}
    \begin{tabular}{lcc}
    \toprule
        Protocol & SMPL annotation & Valid frames \\
    \midrule
        \#PH (HMMR~\cite{DBLP:conf/cvpr/KanazawaZFM19})   & The fits (neutral)        & 26234   \\
        \#PV (VIBE~\cite{kocabas2020vibe})   & Original            & 34561   \\
        \#PS (SPIN~\cite{kolotouros2019learning})   & Original            & 35515   \\
    \bottomrule
    \end{tabular}
    \label{tab:setting_diff}
\end{table}


\begin{table}[t]
\centering
\caption{Results on the 3DPW test set using three data preprocessing protocols. \textbf{First block:} End-to-end approaches. $\dagger$ denotes models that are trained on the 3DPW training split. $\ast$ denotes models that are fine-tuned on the 3DPW test set under 2D supervisions. \textbf{Second block:} Optimization-based approaches that tune the SMPL parameters on the target domain in an offline manner under the supervision of 2D keypoints.
}
\vspace{-5pt}
    \begin{tabular}{lcccc}
    \toprule 
        Method                                               &Prot.         &PA-MPJPE$^\downarrow$ &MPJPE$^\downarrow$ &PVE$^\downarrow$    \\
    \midrule
        HMR~\cite{kanazawa2018end}                              &\#PH       &76.7                   &130.0 &-  \\
        Sim2Real~\cite{DBLP:conf/nips/DoerschZ19}               &\#PH       &74.7                   &-      \\
        HMMR~\cite{DBLP:conf/cvpr/KanazawaZFM19}                &\#PH       &72.6                   &116.5  &- \\
        $^\dagger$VIBE~\cite{kocabas2020vibe}                       &\#PV       &51.9                   &82.9   & 99.1\\
        $^\dagger$ROMP(HRNet-32)~\cite{sun2020monocular}            &\#PV       &47.3                   &76.7 &93.4 \\
        GraphCMR~\cite{kolotouros2019convolutional}             &\#PS       &70.2                   &-  &-    \\
        SPIN~\cite{kolotouros2019learning}                      &\#PS       &59.2                   &96.9 &135.1   \\
        PyMAF~\cite{pymaf2021}                                  &\#PS       &58.9                   &92.8 &110.1  \\
        I2L-MeshNet~\cite{Moon_2020_ECCV_I2L-MeshNet}           &\#PS       &58.6                   &93.2  &- \\
        DaNet(HRNet-48)~\cite{zhang2020learning}                &\#PS       &54.8                   &85.5   &110.8\\
        HybrIK~\cite{li2021hybrik}                              &\#PS       &48.8                   &80.0 &94.5 \\
        $^\dagger$METRO~\cite{lin2021end}                           &\#PS       &47.9                   &77.1  &88.2 \\
        $^\dagger$PARE(HRNet-32)~\cite{kocabas2021pare}             &\#PS       &46.4                   &79.1 &94.2 \\
        $^\dagger$Mesh Graphormer~\cite{lin2021-mesh-graphormer}    &\#PS       &45.6                   &74.7  &87.7 \\
        $^\ast$SPIN~\cite{kolotouros2019learning}                &\#PS       &46.0                   &71.4   &97.0   \\
    \midrule
        SMPLify~\cite{bogo2016keep}                             &\#PH       &106.1                  &199.2 &- \\
        Arnab~\etal~\cite{arnab2019exploiting}                  &\#PH       &72.2                   &-    &-  \\
        ISO~\cite{zhang2020inference}                           &\#PS       &70.8                   &-  &-\\
        Song~\etal~\cite{song2020human}                         &\#PS          &55.9                   &- &- \\
        EFT~\cite{joo2020eft}                                   &\#PS       &54.2                   &-      &-\\
    \midrule
        \multirow{3}{*}{Base model}  &\#PH &120.8                  &276.7 &307.8 \\
        &\#PV       &123.8                  &234.4 &257.2     \\
        &\#PS       &123.4                  &230.3 &253.4     \\
    \midrule
        \multirow{3}{*}{DynaBOA} 
        &\#PH       &\textbf{44.4}                   &\textbf{77.4} &\textbf{89.0}     \\
        &\#PV       &\textbf{42.6}                  &\textbf{69.0} &\textbf{84.4}     \\
        &\#PS       &\textbf{40.4} &\textbf{65.5} &\textbf{82.0}     \\
\bottomrule
\end{tabular}
\label{tab:3dpw}
\end{table}

\subsection{Quantitative Results}
\subsubsection{3DPW}

We compare DynaBOA with end-to-end approaches with frame-based losses~\cite{kanazawa2018end,DBLP:conf/nips/DoerschZ19,kolotouros2019convolutional,kolotouros2019learning,pymaf2021,Moon_2020_ECCV_I2L-MeshNet,zhang2020learning,li2021hybrik,lin2021end,lin2021-mesh-graphormer,kocabas2021pare} and temporal losses~\cite{DBLP:conf/cvpr/KanazawaZFM19,kocabas2020vibe,sun2020monocular,sun2019human}.
We also include iterative optimization-based methods~\cite{bogo2016keep,arnab2019exploiting,joo2020eft,zhang2020inference,song2020human} into comparison.
The evaluation results are presented in \tab{tab:3dpw}, where the second column is the protocol used in the original literature of each compared method.

Compared with the end-to-end methods in \tab{tab:3dpw} (the first block), DynaBOA achieves better performance under all evaluation protocols and particularly outperforms the methods with a carefully designed training scheme to improve the generalization ability of the model~\cite{DBLP:conf/nips/DoerschZ19,sun2019human,zhang2020learning}. 
Notably, DynaBOA also outperforms the methods trained directly on the training set of 3DPW~\cite{kocabas2020vibe,sun2020monocular,lin2021end,lin2021-mesh-graphormer,kocabas2021pare}. 
Furthermore, we provide stronger competitors, \ie, $^\ast$SPIN, by fine-tuning SPIN\footnote{SPIN is a typical method that unifies an optimization module (\eg, SMPLify~\cite{bogo2016keep}) and a neural network in the training framework.} on the test set of 3DPW. We fine-tune the officially released models for $5$ epochs using their source code and hyper-parameter setting. For a fair comparison, only 2D keypoints are available during fine-tuning.
%
DynaBOA significantly outperforms $^\ast$SPIN, which indicates that our test-time adaptation approach can better mitigate the distribution shift by properly exploiting the streaming data from the test domain. 


Moreover, DynaBOA shows its superiority in PA-MPJPE to the optimization-based methods~\cite{bogo2016keep,arnab2019exploiting,zhang2020inference,song2020human,joo2020eft} in \tab{tab:3dpw} (the second block), which iteratively adapt the model on the test set of 3DPW in an offline manner. 
That is to say, the neural networks or the SMPL parameters are optimized independently on each target image.
Compared with these methods, DynaBOA continuously refines the model on the new domain, where the dynamic bilevel optimization scheme, the motion losses, as well as the mean-teacher constraints, all benefit the learning process on the non-stationary distribution of the streaming data. 

To better understand the improvement of DynaBOA, we also report the results of the base model $\mathcal{M}_{\phi_0}$ in \tab{tab:3dpw} (the third block), which is trained on the source dataset only. Note that DynaBOA remarkably reduces the PVE metric by $71.1\%$ on protocol \#PH (from $307.8$mm to $89.0$mm), $67.2\%$ on protocol \#PV and $67.6\%$ on protocol \#PS, which further validates the superiority of dynamic bilevel online adaptation in mitigating domain gaps.

\begin{table}[t]
    \centering
    \caption{Results on the test split of 3DHP. In the second column, some models are shown to be trained on the 3DHP training split ($\mathcal{D}^\text{train}$), while other models are trained purely on the source dataset but with special designs to improve the generalization ability.
    }
    \vspace{-5pt}
    \begin{tabular}{l@{\hspace{0.05in}}c@{\hspace{0.05in}}c@{\hspace{0.05in}}c@{\hspace{0.05in}}@{\hspace{0.05in}}c@{\hspace{0.05in}}c@{\hspace{0.05in}}c}
    \toprule
        Method                                     & $\mathcal{D}^\text{train}$   &PA-MPJPE$^\downarrow$           &MPJPE$^\downarrow$ &PCK$^\uparrow$ &AUC$^\uparrow$  \\
    \midrule
        Mehta~\etal~\cite{mehta2017monocular}      &\Checkmark                         &-                               &-              &72.5           &36.9           \\
        Vnect~\cite{mehta2017vnect}                &\Checkmark                         &98.0                            &124.7              &76.6           &40.4           \\
        EpipolarPose~\cite{kocabas2019self}        &\Checkmark                         &-                               &109.0              &77.5           &-              \\
         HMR~\cite{kanazawa2018end}                                        &\Checkmark                         &89.8                            &124.2              &72.9           &36.5\\
        SPIN~\cite{kolotouros2019learning}                                       &\Checkmark                         &67.5                            &105.2              &76.4           &37.1\\
        \midrule
        Zhou~\etal~\cite{zhou2017towards}         &\XSolidBrush                        &-                               &137.1              &69.2           &32.5              \\
        Habibie~\etal~\cite{habibie2019wild}       &\XSolidBrush                       &92.0                               &127.0              &69.6           &35.5           \\
        \midrule
         Base model                                   &\XSolidBrush                     &116.4                           &199.7              &75.6              &38.4 \\
         DynaBOA                                        &\XSolidBrush                  &\textbf{66.1}                   &\textbf{101.5}     &\textbf{79.5}     &\textbf{43.1}\\
    \bottomrule
    \end{tabular}
    \label{tab:3dhp}
\end{table}

\begin{table}[t]
    \centering
    \caption{Results on the SURREAL test split ($\mathcal{D}^\text{test}$). \textbf{First block:} models are trained with no access to the SURREAL data. \textbf{Second block:} models are trained on the SURREAL training split ($\mathcal{D}^\text{train}$). \textbf{Third block:} models are trained on $\mathcal{D}^\text{test}$ with 2D supervisions. In particular, $^\ast$DecoMR is trained with 3D supervisions on $\mathcal{D}^\text{train}$.
    }
    \vspace{-5pt}
    \begin{tabular}{lccccc}
    \toprule
         Method                             &$\mathcal{D}^\text{train}$     &$\mathcal{D}^\text{test}$   &PA-MPJPE$^\downarrow$    &MPJPE$^\downarrow$ &PVE$^\downarrow$ \\
    \midrule
         SPIN~\cite{kolotouros2019learning}                               &\XSolidBrush                            &\XSolidBrush                          &68.6                       &105.6                  &133.7                 \\GraphCMR ~\cite{kolotouros2019convolutional}                          &\XSolidBrush                            &\XSolidBrush                          &72.0                       &127.4                  &72.0                  \\
         
         DecoMR~\cite{zeng20203d}           &\XSolidBrush                            &\XSolidBrush                          &74.9                       &138.0                  &161.0              \\
         \midrule
         $^\dagger$SPIN                         &\Checkmark                              &\XSolidBrush                          &43.7                       &66.7                   &82.3                 \\
         $^\dagger$GraphCMR                     &\Checkmark                              &\XSolidBrush                          &63.2                       &87.4                   &103.2                  \\ 
         $^\dagger$DecoMR     &\Checkmark                              &\XSolidBrush                          &\underline{43.0}                       &\textbf{52.0}                   &\textbf{68.9}              \\
         \midrule
         $^\ast$SPIN                         &\XSolidBrush                            &\Checkmark                            &51.9                       &80.7                   &99.3                 \\
         $^\ast$GraphCMR                     &\XSolidBrush                            &\Checkmark                            &89.4                       &167.2                         &189.2                 \\
         $^\ast$DecoMR     &\XSolidBrush                            &\Checkmark                            &116.4                       &175.8                   &227.4              \\
         \midrule
         Base model                           &\XSolidBrush                            &\XSolidBrush                        &143.1                      &258.0                   &278.6             \\
         DynaBOA                            &\XSolidBrush                            &\Checkmark                            &\textbf{34.0}                       &\underline{55.2}                   &\underline{70.7}         \\
    \bottomrule
    \end{tabular}
    \label{tab:surreal}
\end{table}

\subsubsection{3DHP} 
We then quantitatively compare DynaBOA with various state-of-the-art methods on 3DHP, including the methods trained on the 3DHP training split~\cite{mehta2017monocular,mehta2017vnect,kocabas2019self,kanazawa2018end,kolotouros2019learning}, trained with multi-view supervision~\cite{kocabas2019self}, and trained with weak supervision signal~\cite{zhou2017towards,habibie2019wild}.
Besides, we report the performance of our base model, which is trained on Human3.6M. The results are reported in \tab{tab:3dhp}. 
We can see that the dynamic online adaptation significantly improves the performance in all metrics. 
After all, as mentioned in \sect{sec:dataset}, the test set of 3DHP only has $2{,}929$ valid images, which indicates that although the target streaming video is very short, DynaBOA can quickly correct the bias of the base model and improve the performance on the unseen target domain.

\subsubsection{SURREAL}
As a large-scale synthetic dataset, SURREAL provides a more significant distribution shift from Human3.6M (see \tab{tab:dataset_gap}). On this dataset, we first compare DynaBOA with SPIN, a representative SMPL-based method. 
We also include in the comparison two typical SMPL-free methods: GraphCMR, which treats vertices as graph nodes, and DecoMR, which predicts UV position images.

The compared models are trained under three different setups, with the results shown in different blocks of \tab{tab:surreal} (from top to bottom).
In Lines 1-3, models are trained on Human3.6M and other data sources excluding the target dataset. We thus directly evaluate the officially released models of the compared methods on the test set of SURREAL. 
In Lines 4-6, we fine-tune the released models of GraphCMR and SPIN \textbf{on the training split of SURREAL} for $5$ epochs under supervisions of 2D keypoints. For DecoMR, we directly use its released model that is trained on the SURREAL training data with 3D annotations.
In Lines 7-9, similar to the idea of DynaBOA, we fine-tune the above models \textbf{on the test split of SURREAL} in an offline manner, also under the supervision of 2D keypoints. 
Obviously, DynaBOA deals with a more challenging problem because it is more difficult to estimate the target distribution from a single frame in the streaming data than from batches of target frames.

From \tab{tab:surreal}, we have two observations: First, models trained on the SURREAL training set (Lines 4-6) obtain remarkable improvements over those trained without SURREAL data (Lines 1-3), which indicates that the distribution shift between source and target domains has significant impacts on the performance. 
Second, DynaBOA achieves the best PA-MPJPE result, and the second-best results in the other two metrics (slightly worse than $^\dagger$DecoMR that is trained with 3D supervisions in SURREAL). 
Notably, $^\dagger$DecoMR outperforms $^\ast$DecoMR by a large margin, which validates the impact of using target domain 3D supervisions. 
Despite the absence of these 3D annotations, DynaBOA effectively reduces the distribution shift through space-time constraints, source exemplar guidance, as well as the bilevel optimization scheme for a compatible use of multi-objectives.


\begin{table}[t]
    \centering
    \caption{Results of using different online optimization schemes (\#PS on 3DPW). ``Single'' indicates the single-level multi-objective optimization, in which B4-B6 optimize multi-objectives in two stages. ``$\rightarrow$'' indicates utilizing the losses on the left first and then on the right in different optimization steps.
    }
    \vspace{-5pt}
    \resizebox{\linewidth}{!}{
    \begin{tabular}{lllccc}
    \toprule
    Model   &Optim.  &Losses    &PA-MPJPE   &MPJPE &PVE  \\
    \midrule
    {B1}         &Single &$\mathcal{L}_{T}$ &114.9      &209.9  &231.3  \\    
    {B2}         &Single &$\mathcal{L}_{F}$ &54.0       &79.5   &108.0\\
    {B3}         &Single &$\mathcal{L}_{F}+\mathcal{L}_{T}$ &55.1       &107.9   &117.8\\ 
    \midrule
    {B4}         &Single &$\mathcal{L}_{F}\rightarrow \mathcal{L}_{T}$ &221.2      &372.0  &505.5\\   
    {B5}         &Single &$\mathcal{L}_{F}\rightarrow \mathcal{L}_{F}$ &58.0       &90.5   &118.4\\
    {B6}         &Single &$\mathcal{L}_{F}\rightarrow \mathcal{L}_{F}+\mathcal{L}_{T}$ &50.1       &78.2   &101.9\\  
    \midrule
    {B7}         &Bilevel  &$\mathcal{L}_{F}\rightarrow \mathcal{L}_{T}$                      &336.7        &502.8   &647.2\\   
    {B8}         &Bilevel                        &$\mathcal{L}_{F}\rightarrow \mathcal{L}_{F}$                      &49.0      &75.2  &100.0\\   
    {BOA}        &Bilevel                        &$\mathcal{L}_{F}\rightarrow \mathcal{L}_{F}+\mathcal{L}_{T}$      &\textbf{48.1}       &\textbf{73.5}   &\textbf{96.1} \\   
    \bottomrule
    \end{tabular}
    }
    \label{tab:bilevel_ablates}
\end{table}

\subsection{Qualitative Results}

\begin{figure*}[!t]
    \centering
    \includegraphics[width=\textwidth]{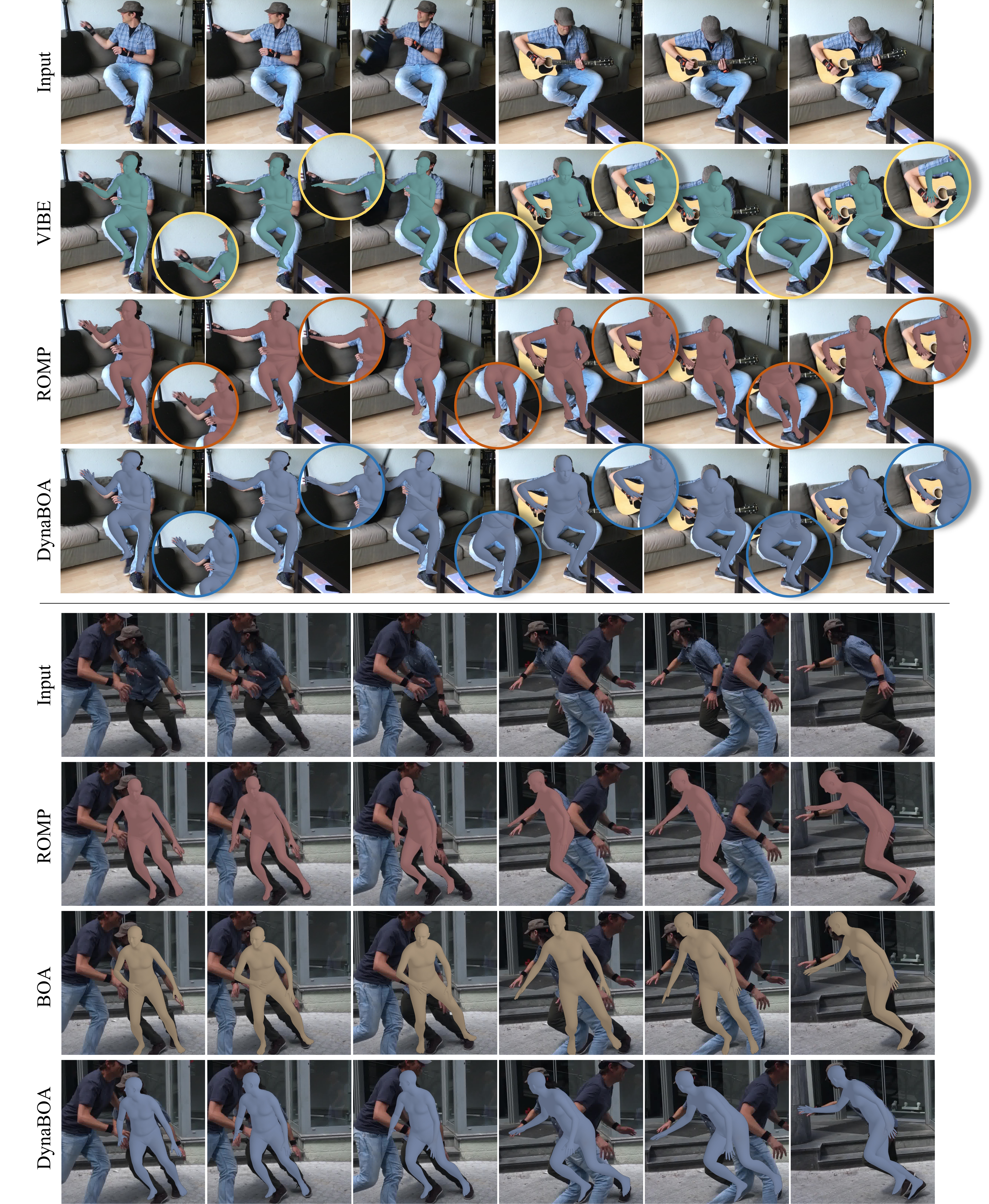}
    \vspace{-15pt}
        \caption{\textbf{Top:} Qualitative showcases of mesh reconstruction on 3DPW streaming data, where the action of playing guitar is not included in the Human3.6M source domain. Note that VIBE and ROMP are specifically designed for videos, and they take the training data of 3DPW (along with the 3D annotations) as a part of the training set. We zoom in on the limbs for better visualization. \textbf{Bottom:} An example of mesh reconstruction under severe occlusion, in which the subject of interest (\ie, the man behind) is heavily occluded by another man. It is an extremely out-of-domain scenario because the source domain does not contain any occlusion samples.}
    \label{fig:comparison}
\end{figure*}

In \fig{fig:comparison} (Top), we present a typical showcase of mesh reconstruction on the challenging 3DPW dataset. We here select a sequence of playing guitar as the example, which is extremely different from the actions in Human3.6M. 
The compared methods, including VIBE and ROMP, are also specifically designed for videos and take the training data of 3DPW (along with the 3D annotations) as a part of the training set. 
However, in this example, the compared models cannot correctly estimate the positions of the arms and legs. In contrast, our model can better align the human meshes to the ground-truth body structures. 
This verifies that DynaBOA can effectively adapt the model to novel actions in streaming data.

To demonstrate that DynaBOA contributes to the adaptation to difficult key-frames, in \fig{fig:comparison} (Bottom), we show the mesh reconstruction results of target frames under severe occlusions. We can see that the subject of interest (the man behind) is heavily occluded by another man, especially in the middle two frames.
It represents an extremely out-of-domain scenario, as the source domain of Human3.6M does not contain any occlusion samples. 
Notably, both ROMP (trained on the target training set with 3D annotations) and the original BOA~\cite{guan2021bilevel} (without the new methods of exemplar guidance and dynamic update) fail to estimate reasonable postures of the occluded man. 
In contrast, DynaBOA is the only one that successfully survives in this hard case with extraordinary domain gaps. 
DynaBOA also gives more accurate estimations of head orientation than the compared methods. It is an interesting result, because no 2D head keypoints, \eg, eyes and nose, are used in the online adaptation phase. A possible reason is that DynaBOA effectively transfers the prior knowledge about the 3D postures from the retrieved source exemplar. 


\begin{table}[t]
    \centering
    \caption{Ablation studies on different forms of temporal constraints used in the upper-level BOA step (\#PS on 3DPW).}
    \vspace{-5pt}
    \begin{tabular}{lccccc}
    \toprule
         Model                          & $\mathcal{L}_\text{motion}$     & $\mathcal{L}_\text{teacher}$      &PA-MPJPE            &MPJPE        &PVE          \\
    \midrule
         B8                  &\XSolidBrush                       &\XSolidBrush                           &49.0            &75.2         &100.0            \\
        B9                    &\Checkmark                         &\XSolidBrush                           &48.2            &73.7         &98.2            \\
         B10                    &\XSolidBrush                       &\Checkmark                             &48.7            &74.9         &97.9            \\
         {BOA}                 &\Checkmark                         &\Checkmark                               &\textbf{48.1}          &\textbf{73.5}        &\textbf{96.1}           \\
    \bottomrule 
    \end{tabular}
    \label{tab:temporal_ablates}
\end{table}

\subsection{Analyses of the Bilevel Optimization Scheme}

\myparagraph{Alternative online optimization schemes.} 
Previously in \sect{sec:boa}, we discuss the straightforward single-level optimization schemes of online adaptation.
Table~\ref{tab:bilevel_ablates} gives the corresponding results obtained under the \#PS protocol on 3DPW.
We have the following observations.
First, the space-time multi-objectives are beneficial with carefully designed optimization schemes, as B6 and BOA outperform B5 and B8 accordingly.
However, as B3 performs worse than B2, we may conclude that a simple combination of the multi-objectives easily leads to sub-optimal results. 
%
Second, BOA achieves consistent improvements over B6 in all three evaluation metrics, which indicates that the bilevel optimization scheme can effectively combine the best of both frame-wise and temporal constraints. 
The major difference between BOA and B6 is that whether the second training step under $\mathcal{L}_F+\mathcal{L}_T$ is to update the parameters in $\mathcal{M}_{\widetilde{\boldsymbol{\phi}}_{t}}$ (in BOA) or $\mathcal{M}_{\phi_{t-1}}$ (in B6).

\begin{figure}[t]
    \centering
    \includegraphics[width=\linewidth]{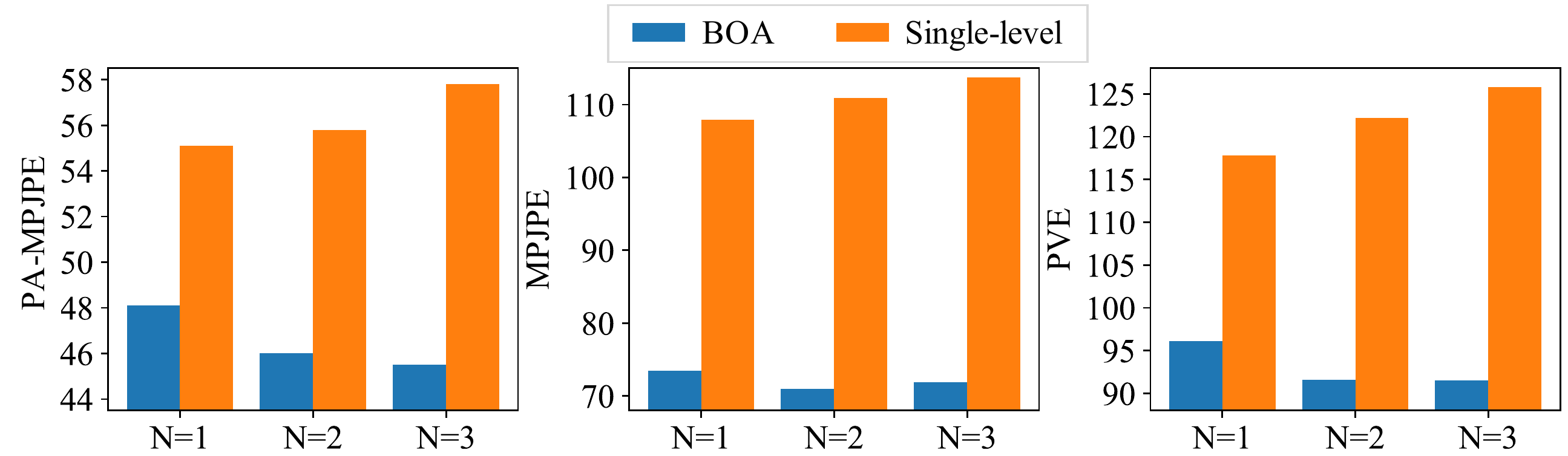}
    \vspace{-15pt}
    \caption{Results of optimizing the model $N$ steps on each frame (\ie, repeating the operations in \underline{Lines 4-9} in \alg{alg:BOA} for $N$ times). As $N$ grows, BOA performs better than its single-level counterpart, showing the ability to prevent overfitting. Experiments are conducted under the \#PS protocol on 3DPW.
    }
    \label{fig:overfit_ablates}
\end{figure}

\begin{figure}[t]
    \centering
    \includegraphics[width=0.95\linewidth]{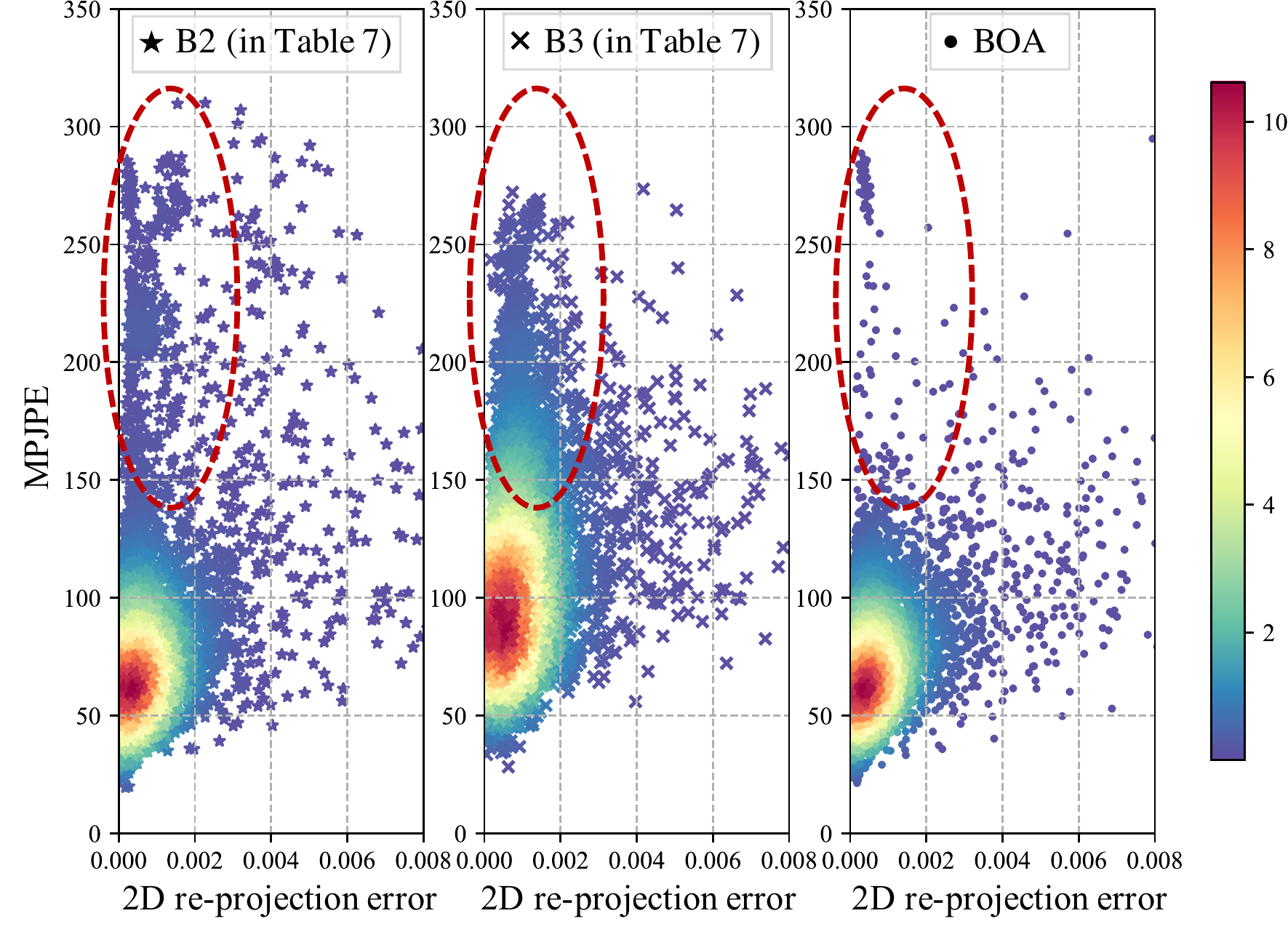}
    \vspace{-5pt}
    \caption{Correlations between the 2D re-projection error and the evaluation metric. Each point represents a target sample. We use the color intensity to indicate the point density. The points inside the dashed ellipses are typical samples suffering from 3D ambiguity.
    }
    \label{fig:loss-metric}
\end{figure}

\myparagraph{Different forms of temporal constraints.} 
Table~\ref{tab:temporal_ablates} shows the ablation studies for the two components of the proposed temporal constraints, \ie, $\mathcal{L}_\text{motion}$ and $\mathcal{L}_\text{teacher}$, that are used in the upper-level optimization step of BOA. 
We observe that, first, the individual use of $\mathcal{L}_\text{motion}$ (B9) and $\mathcal{L}_\text{teacher}$ (B10) respectively improves the final results of the baseline model only trained with frame-wise losses (B8).
Second, the advantages of these two loss components are complementary, as BOA consistently outperforms the two baseline models (B9 and B10).
A possible reason is that $\mathcal{L}_\text{motion}$ captures short-term temporal consistency of shapes and postures (in a short video snippet with a length of $\tau$), while $\mathcal{L}_\text{teacher}$ focuses more on long-term information.

\myparagraph{BOA vs. overfitting.} 
As we have discussed in \sect{sec:boa}, the bilevel optimization can facilitate the compatibility of multi-objectives and prevent the model from overfitting to either $\mathcal{L}_F$ or $\mathcal{L}_T$.  
To verify this, we increase the number of optimization steps on each test image, \ie, repeating the operations in \underline{Lines 4-9} in \alg{alg:BOA} for $N$ times, and compare the results with a single-level optimization baseline (B3 in \tab{tab:bilevel_ablates}), which trains the model with an objective function of $\mathcal{L}_F+\mathcal{L}_T$.
As shown in \fig{fig:overfit_ablates}, with the growth of the optimization steps, the error of the single-level optimization scheme increases quickly in all metrics including PA-MPJPE, MPJPE, and PVE. It indicates that a direct combination of the space-time constraints is prone to result in overfitting to the current video frame, and thus makes it difficult for the model to quickly adapt to the next frame. 
In contrast, as $N$ grows, BOA achieves better performance, indicating that the proposed bilevel optimization scheme can alleviate overfitting.

\myparagraph{Can BOA reduce 3D ambiguity?}
In \fig{fig:loss-metric}, the X-axis refers to the 2D keypoint re-projection error the Y-axis is the 3D mesh reconstruction error in MPJPE. Each point corresponds to a target sample, and we estimate the probability density of target samples using Gaussian kernels. 
The hot spot indicates areas with higher point density. 
The three sub-figures respectively show the results of the vanilla baseline only trained with frame-wise loss functions (B2 in \tab{tab:bilevel_ablates}), the results of the baseline model trained with multi-objectives ({B3}), and those of the model trained with BOA.
As shown by the point density inside the dashed areas (with low 2D error but high 3D error), both B2 and B3 suffer from the problem of 3D ambiguity, while BOA effectively alleviates this issue due to effective optimization of the space-time constraints. 



\subsection{Analyses of New Contributions in DynaBOA}

We provide in-depth ablation studies on the new components in DynaBOA based on its predecessor~\cite{guan2021bilevel}, \ie, 3D exemplar guidance and dynamic update. We include more experiments about hyperparameter sensitivity in \underline{Supplementary Materials}.

\myparagraph{Ablation studies.} 
We make two extensions on the BOA: the 3D exemplar guidance and the dynamic update strategy. \tab{tab:ablation_constributions} gives the ablation results of each component, from which we observe that individually performing cross-domain retrieval or dynamic update gains significant improvements over the original BOA model. 
For example, B11 reduces PA-MPJPE by $2.5$mm and B12 reduces PA-MPJPE by $4.1$mm.
Furthermore, by jointly using these two techniques in DynaBOA, we obtain more performance gains, reducing PA-MPJPE by $7.7$mm. 
The above results suggest that the advantages of the new components are complementary, which is in line with our expectations, as the exemplar guidance is designed to compensate for the lack of direct 3D supervisions, while the dynamic update strategy is designed to enhance online adaptation on difficult streaming samples.

\begin{table}[t]
    \centering
    \caption{Ablation studies on the new contributions of DynaBOA based on BOA (\#PS on 3DPW). ``Exemplar'' refers to the exemplar guidance method. ``Dynamic'' refers to the dynamic update strategy. 
    }
    \vspace{-5pt}
    \begin{tabular}{lccccc}
    \toprule
         Model             & Exemplar   & Dynamic &PA-MPJPE &MPJPE &PVE \\
    \midrule
        {BOA}       &\XSolidBrush   &\XSolidBrush           &48.1 &73.6 &96.1  \\
         {B11}       &\Checkmark     &\XSolidBrush           &45.6 &73.1 &90.2  \\
         {B12}       &\XSolidBrush   &\Checkmark             &44.0 &67.6 &89.2  \\
         {DynaBOA}   &\Checkmark     &\Checkmark             &\textbf{40.4} &\textbf{65.5} &\textbf{82.0}  \\
    \bottomrule
    \end{tabular}
    \label{tab:ablation_constributions}
\end{table}

\begin{figure}[t]
    \centering
    \includegraphics[width=\linewidth]{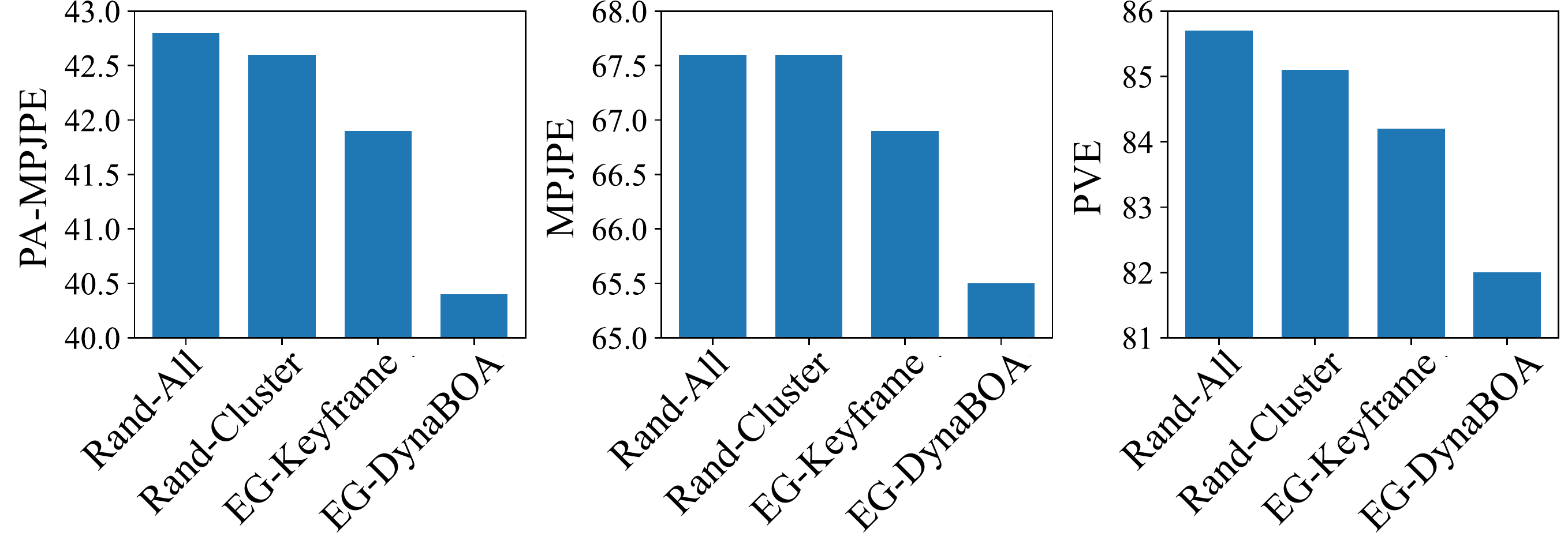}
    \vspace{-20pt}
    \caption{Comparison of reconstruction errors (\#PS protocol on 3DPW) of different retrieval schemes for 3D exemplar guidance. See text for detailed descriptions of each scheme.
    }
    \label{fig:retrieval_ablates}
\end{figure}

\myparagraph{Analyses on 3D exemplar guidance.} 
Compared to the previous conference version, we replace the random selection of source exemplars with efficient retrieval.
To examine this modification, we take two random selection schemes as baselines: One is random selection from the whole source dataset (``Rand-All'' in \fig{fig:retrieval_ablates}) and the other is random selection from the source clusters $\{\mathcal{S}^{\boldsymbol{c}_{n}}\}_{n=1}^{N_C}$ (see \fig{fig:method_retrieval}) that are preprocessed offline (``Rand-Cluster''). 
In addition, one may consider what if the proposed retrieval process is only performed on the key-frames that have more than one model refinement step in the process of dynamic update (``EG-Keyframe''). 
From \fig{fig:retrieval_ablates}, we observe that random selection schemes perform much worse than the approaches based on similarity-driven retrieval. 
The reason is that random selection would inevitably bring in source samples with a large distribution shift from the current target sample, thus playing a negative role in online adaptation.
By comparing ``EG-Keyframe'' with ``EG-DynaBOA'', which is the final scheme in this work, we find the necessity of retrieving source exemplars for every target frame, even including the regular target frames.

\begin{figure*}[t]
    \centering
    \includegraphics[width=\linewidth]{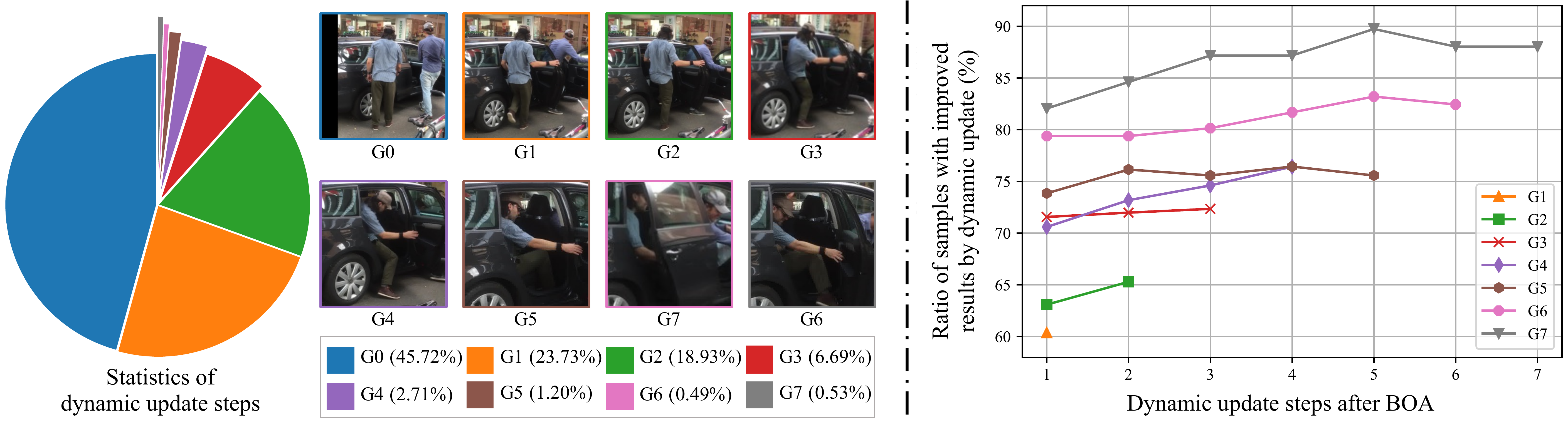}
    \vspace{-15pt}
    \caption{\textbf{Left:} The ratio of test frames with different steps of model refinement in the process of dynamic update and corresponding showcases from 3DPW. \textbf{Right:} The ratio of samples in the groups of G1--G7 with reconstruction results after $l$ steps of model refinement.
    }
    \label{fig:ana_dynamic_update}
\end{figure*}

\myparagraph{Analyses on the dynamic update strategy.} 
We illustrate the advantage of the dynamic update by counting the steps of model refinement after BOA at each target frame. 
For example, G0 represents the cases where the model is refined $0$ steps after the optimization process of BOA.
From the pie chart, we observe that $45.72\%$ images (G0) have been sufficiently processed by BOA and need no further training steps, while only $4.93\%$ images require more than $3$ additional training steps.
We also show typical target frames corresponding to different numbers of dynamic updates, indicating that the dynamic update strategy allows the model to be trained with an adaptive number of iterations according to the difficulty of each target frame or the degree of variations between consecutive frames.
%
In \fig{fig:ana_dynamic_update} (Right), we show the ratio of samples ($\%$) in the groups of G1--G7 with better reconstruction results after $l$ steps of model refinement, compared with $l=0$.
The upward trend of each curve indicates that the dynamic update strategy is effective for target images in all groups.

\begin{figure}[t]
    \centering
    \includegraphics[width=\linewidth]{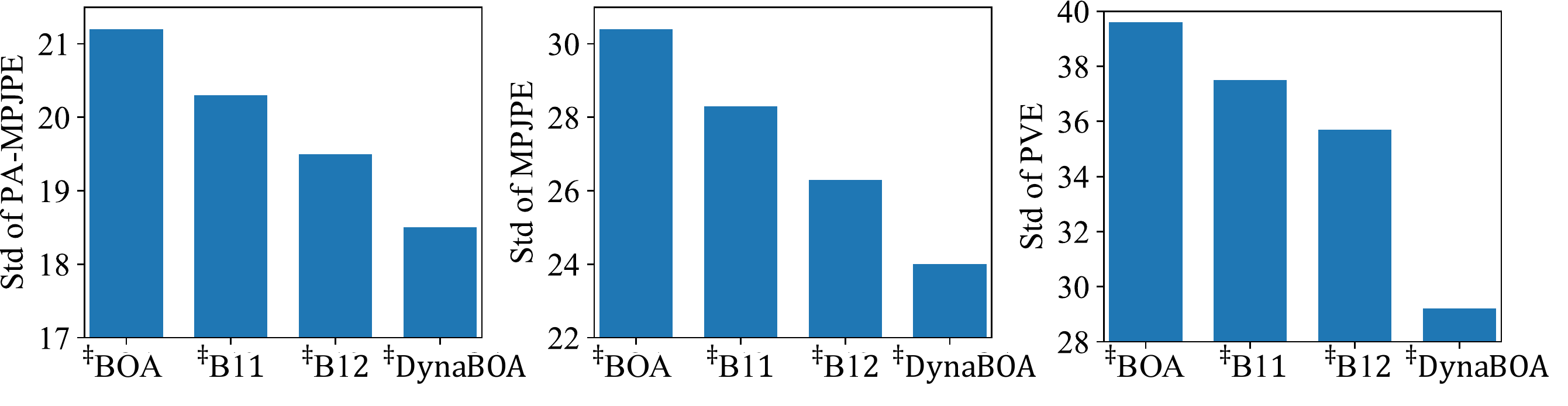}
    \vspace{-15pt}
    \caption{The standard deviations of the reconstruction errors of 3DPW test samples. The superscript $\ddagger$ indicates adapting the models to non-stationary streaming data with highly mixed 3DPW and 3DHP videos, which is different from the standard setting (\tab{tab:ablation_constributions}). Smaller standard deviations of $^\ddagger$DynaBOA indicate being capable of coping with difficult key-frames.
    }
    \label{fig:mixdataset}
\end{figure}

\myparagraph{Adapting to non-stationary streaming data with highly mixed 3DPW and 3DHP videos.} 
To provide the test streaming data with more non-stationary changes, we mix the videos of 3DPW and 3DHP to form a new target dataset. In this case, the scene changes frequently, and the corresponding frames can be regarded as the difficult key-frame.
%
%
%
\fig{fig:mixdataset} gives the standard deviations of the reconstruction errors of 3DPW samples. The smaller standard deviation of $^\ddagger$DynaBOA implies a more stable adaptation process and that our approach can ease the training difficulty at unlabeled key-frames with dramatic scene changes. It finally achieves $41.1$mm / $66.8$mm / $70.4$m in PA-MPJPE / MPJPE / PVE, which are comparable with the results of DynaBOA in \tab{tab:ablation_constributions}.
Note that these two models are trained on different streaming sequences, but evaluated on the same set of 3DPW.

\section{Conclusion}
In this paper, we presented a new research problem of reconstructing human meshes from out-of-domain streaming videos. 
To tackle the distribution shift, we proposed a new test-time adaptation algorithm named DynaBOA. 
The basic idea is to effectively leverage temporal consistency through bilevel online adaptation (BOA), which has been shown to avoid overfitting the ambiguous 2D supervisions.
Besides, we proposed to retrieve source images efficiently as exemplars guidance to further compensate for the lack of 3D annotations in the process of test-time training. 
Furthermore, we introduced the dynamic update strategy, a natural extension of BOA, to enhance the adaptation ability on key-frames while preventing the model from overfitting regular frames.
We conducted experiments on three benchmarks with streaming data that has a remarkable distribution shift from the source domain. 
On all benchmarks, DynaBOA consistently outperforms the state-of-the-art mesh reconstruction methods in the new out-of-domain setup, showing the ability to alleviate the training difficulty when adapting the model to highly non-stationary streaming videos.


%



\ifCLASSOPTIONcompsoc
  \section*{Acknowledgments}
\else
  \section*{Acknowledgment}
\fi

This work was funded by NSFC (U19B2035, U20B2072, 61976137, 62106144), the Shanghai Municipal Science and Technology Major Project (2021SHZDZX0102), and the Shanghai Sailing Program.


\ifCLASSOPTIONcaptionsoff
  \newpage
\fi



%

\bibliographystyle{IEEEtran}
\bibliography{IEEEabrv,boa.bib}
\end{document}